\documentclass{article}
\pdfoutput=1
\PassOptionsToPackage{numbers, compress}{natbib}

\usepackage[preprint]{neurips_2019}

\usepackage[utf8]{inputenc} 
\usepackage[T1]{fontenc}    
\usepackage[hidelinks]{hyperref}       
\usepackage{url}            
\usepackage{booktabs}       
\usepackage{amsfonts}       
\usepackage{nicefrac}       
\usepackage{microtype}      
\usepackage{graphicx}
\usepackage{subfig}
\usepackage[export]{adjustbox}
\usepackage{amsmath, amssymb}
\usepackage{tikz}
\usepackage{pgfplots}

\pgfplotsset{mystyle/.append style={axis x line=middle, axis y line=middle, xlabel={$x$}, ylabel={$y$}, axis equal }}
\tikzset{color dash/.style={dash pattern=on 5pt off 15pt,postaction={draw,red,dash phase=5pt},postaction={draw,green,dash phase=10pt},postaction={draw,blue,dash phase=15pt}}}
\tikzset{short color dash/.style={dash pattern=on 2pt off 6pt,postaction={draw,red,dash phase=2pt},postaction={draw,green,dash phase=4pt},postaction={draw,blue,dash phase=6pt}}}
\usepgfplotslibrary{external,statistics} 

\usepackage{tabularx}
\usetikzlibrary{arrows,automata,calc,backgrounds,decorations.pathreplacing,fit}
\usepackage{chemfig}
\usepackage{verbatim}
\usepackage[normalem]{ulem}
\usepackage{seqsplit}

\newcommand{\KL}{\text{KL}}

\title{All SMILES Variational Autoencoder}

\author{%
  Zaccary Alperstein \\
  Quadrant \\
  \texttt{zac@quadrant.ai}
  \And
  Artem Cherkasov \\
  Vancouver Prostate Centre, UBC \\
  \texttt{acherkasov@prostatecentre.com}
  \And
  Jason Tyler Rolfe \\
  Quadrant \\
  \texttt{jason@quadrant.ai}
}

\begin{document}

\maketitle

\begin{abstract}
  Variational autoencoders~(VAEs) defined over SMILES string and graph-based representations of molecules promise to improve the optimization of molecular properties, thereby revolutionizing the pharmaceuticals and materials industries. However, these VAEs are hindered by the non-unique nature of SMILES strings and the computational cost of graph convolutions. 
To efficiently pass messages along all paths through the molecular graph,
we encode multiple SMILES strings of a single molecule using a set of stacked recurrent neural networks, pooling hidden representations of each atom between SMILES representations, 
and use attentional pooling to build a final fixed-length latent representation.
By then decoding to a disjoint set of SMILES strings of the molecule, our All SMILES VAE learns an almost bijective mapping between molecules and latent representations near the high-probability-mass subspace of the prior. 
Our SMILES-derived but molecule-based latent representations significantly surpass the state-of-the-art in 
a variety of fully- and semi-supervised property regression and molecular property optimization tasks.
\end{abstract}

\section{Introduction}
\label{Introduction}

The design of new pharmaceuticals, OLED materials, and photovoltaics all require optimization within the space of molecules~\cite{pyzer2015high}. While well-known algorithms ranging from gradient descent to the simplex method facilitate efficient optimization, they generally assume a continuous search space and a smooth objective function. In contrast, the space of molecules is discrete and sparse. Molecules correspond to graphs, with each node labeled by one of ninety-eight naturally occurring atoms, and each edge labeled as a single, double, or triple bond. Even within this discrete space, almost all possible combinations of atoms and bonds do not form chemically stable molecules, and so must be excluded from the optimization domain, yet there remain as many as $10^{60}$ small molecules to consider~\cite{reymond2010chemical}. Moreover, properties of interest are often sensitive to even small changes to the molecule~\cite{stumpfe2012exploring}, so their optimization is intrinsically difficult.

Efficient, gradient-based optimization can be performed over the space of molecules given a map between a continuous space, such as $\mathbb{R}^n$ or the $n$-sphere, and the space of molecules and their properties~\cite{sanchez2018inverse}.
Initial approaches of this form trained a variational autoencoder (VAE)~\cite{kingma2013auto, rezende2014stochastic} on SMILES string representations of molecules~\cite{weininger1988smiles} to learn a decoder mapping from a Gaussian prior to the space of SMILES strings~\citep{gomez2018automatic}. A sparse Gaussian process on molecular properties then facilitates Bayesian optimization of molecular properties within the latent space~\cite{dai2018syntax, gomez2018automatic, kusner2017grammar, samanta2018nevae}, or a neural network regressor from the latent space to molecular properties can be used to perform gradient descent on molecular properties with respect to the latent space~\cite{aumentado2018latent, jin2018junction, liu2018constrained, mueller2017sequence}.
Alternatively, semi-supervised VAEs condition the decoder on the molecular properties~\cite{kang2018conditional, lim2018molecular}, so the desired properties can be specified directly.
Recurrent neural networks have also been trained to model SMILES strings directly, and tuned with transfer learning, without an explicit latent space or encoder~\cite{gupta2018generative, segler2017generating}.

\begin{figure}[tbh]
\center
\includegraphics[width=0.95\textwidth]{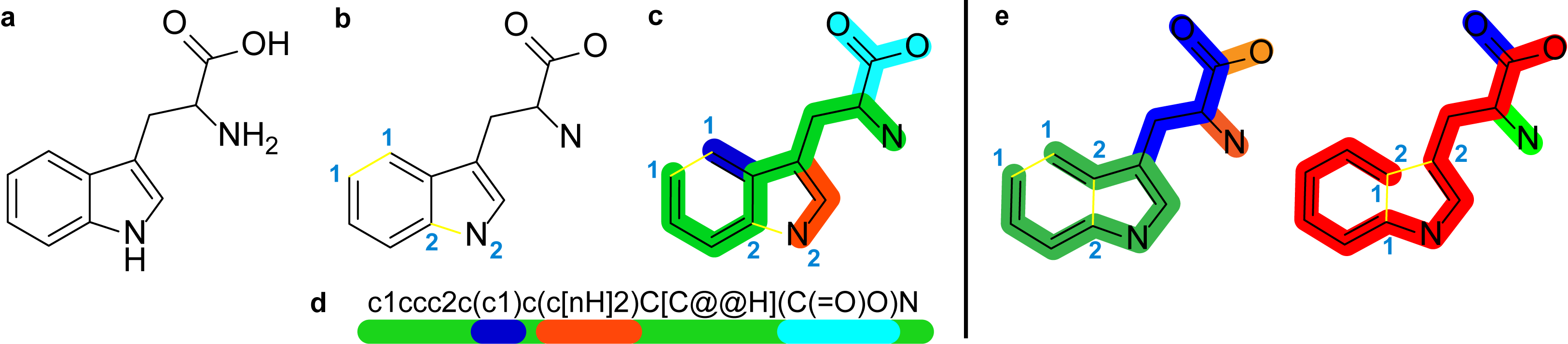}
  \caption{The molecular graph of the amino acid Tryptophan~(a). To construct a SMILES string, all cycles are broken, forming a spanning tree~(b); a depth-first traversal is selected~(c); and this traversal is flattened~(d). The beginning and end of intermediate branches in the traversal are denoted by~( and~) respective. The ends of broken cycles are indicated with matching digits. The full grammar is listed in Appendix~\ref{sec:SMILES_grammar}. A small set of SMILES strings can cover all paths through a molecule (e).}
  \label{fig:SMILES_def}

\end{figure}

SMILES, the simplified molecular-input line-entry system, defines a character string representation of a molecule by performing a depth-first pre-order traversal of a spanning tree of the molecular graph, emitting characters for each atom, bond, tree-traversal decision, and broken cycle~\cite{weininger1988smiles}. The resulting character string corresponds to a flattening of a spanning tree of the molecular graph, as shown in Figure~\ref{fig:SMILES_def}.
The SMILES grammar is restrictive, and most strings over the appropriate character set do not correspond to well-defined molecules. Rather than require the VAE
decoder to explicitly learn this grammar, context-free grammars~\cite{kusner2017grammar}, and attribute grammars~\cite{dai2018syntax} have been used to constrain the decoder,
increasing the percentage of valid SMILES strings produced by the generative model. 
Invalid SMILES strings and violations of simple chemical rules can be avoided entirely by operating on the space of molecular graphs, either directly~\cite{de2018molgan, li2018learning, liu2018constrained, ma2018constrained, simonovsky2018graphvae} or 
via junction trees~\cite{jin2018junction}. 

Every molecule is represented by many well-formed SMILES strings, corresponding to all depth-first traversals of every spanning tree of the molecular graph. The distance between different SMILES strings of the same molecule can be much greater than that between SMILES strings from radically dissimilar molecules~\cite{jin2018junction}, as shown in Figure~\ref{fig:samemol_diffsmi} of Appendix~\ref{sec:datasets}. A generative model of individual SMILES strings will tend to reflect this geometry, complicating the mapping from latent space to molecular properties and creating unnecessary local optima for property optimization~\cite{vinyals2015order}. To address this difficulty, sequence-to-sequence transcoders~\cite{sutskever2014sequence} have been trained to map between different SMILES strings of a single molecule~\cite{bjerrum2018improving, winter2019learning, winter2019efficient}. 

Reinforcement learning, often combined with adversarial methods, has been used to train progressive molecule growth strategies~\cite{guimaraes2017objective, jaques2017sequence, olivecrona2017molecular, putin2018reinforced, 
  you2018graph, zhou2018optimization}. While these approaches have achieved state-of-the-art optimization of simple molecular properties that can be evaluated quickly \emph{in silico},
critic-free techniques generally depend upon property values of algorithm-generated molecules (but see~\cite{de2018molgan, popova2018deep}), and so 
scale poorly to real-world properties requiring time-consuming wet-lab experiments. 

Molecular property optimization would benefit from a generative model that directly captures the geometry of the space of molecular graphs, rather than SMILES strings, but efficiently infers a latent representation sensitive to spatially distributed molecular features. 
To this end, we introduce the All SMILES VAE, which uses
recurrent neural networks (RNNs) on
multiple SMILES strings to implicitly perform efficient message passing along and amongst many flattened spanning trees of the molecular graph in parallel. A fixed-length latent representation is distilled from the variable-length 
RNN output using attentional mechanisms. From this latent representation, the decoder RNN reconstructs a set of SMILES strings disjoint from those input to the encoder, ensuring that the latent representation only captures features of the molecule, rather than its SMILES realization. Simple property regressors jointly trained on this latent representation surpass the state-of-the-art for molecular property prediction, and facilitate exceptional gradient-based molecular property optimization when constrained to the region of prior containing almost all probability. We further demonstrate that the latent representation forms a near-bijection with the space of molecules, and is smooth with respect to molecular properties, facilitating effective optimization.

\section{Efficient molecular encoding with multiple SMILES strings}

A variational autoencoder (VAE) defines a generative model over an observed space~$x$ in terms of a prior distribution over a latent space~$p(z)$ and a conditional likelihood of observed states given the latent configuration $p(x|z)$~\cite{kingma2013auto, rezende2014stochastic}. The true log-likelihood $p(x) = \log \left[ \int_z p(z) p(x|z) \right]$ is intractable, so the evidence lower bound (ELBO), based upon a variational approximation $q(z|x)$ to the posterior distribution, is maximized instead:
$\mathcal{L} = \mathbb{E}_{q(z|x)} \left[ \log p(x | z) \right] - \KL\left[q(z|x) || p(z) \right] .$
The ELBO implicitly defines a stochastic autoencoder, with encoder $q(z|x)$ and decoder $p(x|z)$.

Many effective molecule encoders rely upon 
graph convolutions: local message passing in the molecular graph, between either adjacent nodes or adjacent edges~\cite{duvenaud2015convolutional, kearnes2016molecular, kipf2016semi, li2015gated, lusci2013deep}. 
To maintain permutation symmetry, the signal into each node is a sum of messages from the adjacent nodes, but may be a function of edge type, or attentional mechanisms dependent upon the source and destination nodes~\cite{ryu2018deeply}. This sum of messages is then subject to a linear transformation and a pointwise nonlinearity. Messages are sometimes subject to gating~\cite{li2015gated}, like in
long short-term memories (LSTM)~\cite{hochreiter1997long} and gated recurrent units (GRU)~\cite{cho2014learning}, as detailed in Appendix~\ref{sec:GRU_review}.

Message passing on molecular graphs is analogous to a traditional convolutional neural network applied to images~\cite{krizhevsky2012imagenet, lecun1990handwritten}, with constant-resolution hidden layers~\cite{he2016deep} and two kernels: a $3 \times 3$ average-pooling kernel that sums messages from adjacent pixels (corresponding to adjacent nodes in a molecular graph),
and a trainable $1 \times 1$ kernel that transforms the message from each pixel (node) independently, before a pointwise nonlinearity.
While convolutional networks with such small kernels are now standard in the visual domain, they use hundreds 
of layers to pass information throughout the image and achieve effective receptive fields that span the entire input~\cite{szegedy2016rethinking}.
In contrast, molecule encoders generally use between three and seven rounds of message passing~\cite{duvenaud2015convolutional, gilmer2017neural, jin2018junction, kearnes2016molecular, liu2018constrained, samanta2018nevae, you2018graph}.
This limits the computational cost, since molecule encoders cannot use highly-optimized implementations of spatial 2D convolutions, but each iteration of message passing only propagates information a geodesic distance of one within the molecular graph.\footnote{All-to-all connections allow fast information transfer, but computation is quadratic in graph size~\cite{gilmer2017neural, kearnes2016molecular}. \citet{lusci2013deep} considered a set of DAGs rooted at every atom, with full message propagation in a single pass.} 
In the case of the commonly used dataset of 250,000 drug-like molecules~\cite{gomez2018automatic}, 
information cannot traverse these graphs effectively, as their average diameter 
is 11.1, and their maximum diameter is 24, as shown in Appendix~\ref{sec:datasets}.

Non-local molecular properties, requiring long-range information propagation along the molecular graph, are of practical interest in domains including pharmaceuticals, photovoltaics, and OLEDs.
The pharmacological efficacy of a molecule generally depends upon high binding affinity for a particular receptor or other target, and low binding affinity for other possible targets.
These binding affinities 
are determined by the maximum achievable alignment between the molecule's electromagnetic fields and those of the receptor. 
Changes to the shape or charge distribution in one part of the molecule affect the position and orientation at which it fits best with the receptor, inducing shifts and rotations that alter the binding of other parts of the molecule, and changing the binding affinity~\cite{clayden2001organic}.
Similarly, efficient next-generation OLEDs depend on properties, such as the singlet-triple energy gap, that are directly proportional to the strength of long-range electronic interactions across the molecule~\cite{im2017molecular}.
The latent representation of a VAE can directly capture these non-local, nonlinear properties only if the encoder passes information efficiently across the entire molecular graph.

Analogous to graph convolutions, gated RNNs defined directly on SMILES strings effectively pass messages, via the hidden state, through a flattened spanning tree of the molecular graph (see Figure~\ref{fig:SMILES_def}). The message at each symbol in the string is a weighted sum of the previous message and the current input, followed by a pointwise nonlinearity and subject to gating, as reviewed in Appendix~\ref{sec:GRU_review}. This differs from explicit graph-based message passing in that the molecular graph is flattened into a chain corresponding to a depth-first pre-order traversal of a spanning tree, and the set of adjacent nodes that affect a message only includes the preceding node in this chain.
Rather than updating all messages in parallel, RNNs on SMILES strings move sequentially down the chain, so earlier messages influence all later messages, and information can propagate through all branches of a flattening of a spanning tree in a single pass.
With a well-chosen spanning tree, information can pass the entire width of the molecular graph in a single RNN update. The relationship between RNNs on SMILES strings and graph-based architectures is further explored in Appendix~\ref{sec:extended_model_architecture}.

\section{Model architecture}

To marry the latent space geometry induced by graph convolutions to the information propagation efficiency of RNNs on SMILES strings, the All SMILES encoder combines these architectures. It takes multiple distinct SMILES strings of the same molecule as input, and applies RNNs to them in parallel. This implicitly realizes a representative set of message passing pathways through the molecular graph, corresponding to the depth-first pre-order traversals of the spanning trees underlying the SMILES strings. Between each layer of RNNs, the encoder pools homologous messages between parallel representations, so that information flows along the union of the implicit SMILES pathways.

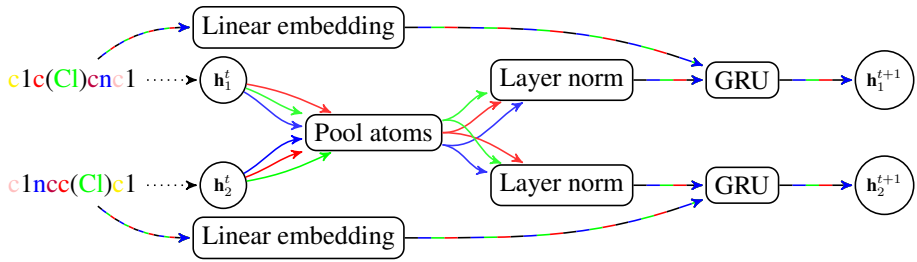
\begin{figure}[tbh]
  \center
  \begin{tikzpicture}[->,>=stealth',shorten >=1pt,auto,node distance=2.0cm,semithick]
    \tikzstyle{every state}=[fill=white,draw=black,text=black,scale=0.7]
    \node[state]     (h11)                   {$\textbf{h}_1^t$};
    \node[state]     (h21) [below of=h11]    {$\textbf{h}_2^t$};
    \node[draw=none] (h11_label) [left of=h11] {{\color{yellow} c}1{\color{red} c}({\color{green}Cl}){\color{purple} c}{\color{blue} n}{\color{pink} c}1};
    \node[draw=none] (h21_label) [left of=h21] {{\color{pink} c}1{\color{blue}n}{\color{purple} c}{\color{red} c}({\color{green} Cl}){\color{yellow} c}1};

    \node[draw=black,rounded corners] (h10_embed) [above of=h11_label,xshift=3.0cm,yshift=-1.3cm] {Linear embedding};
    \node[draw=black,rounded corners] (h20_embed) [below of=h21_label,xshift=3.0cm,yshift=1.3cm]  {Linear embedding};

    \node[draw=none]                      (p_offset_1) [right of=h11]              {};
    \node[draw=none]                      (p_offset_2) [right of=h21]              {};
    \node[draw=black,rounded corners]     (pool)       [right of=h11,yshift=-0.7cm] {Pool atoms};
    \node[draw=black,rounded corners]     (layer_norm_1) [right of=p_offset_1,xshift=0.5cm] {Layer norm};
    \node[draw=black,rounded corners]     (layer_norm_2) [right of=p_offset_2,xshift=0.5cm] {Layer norm};
    \node[draw=black,rounded corners]     (RNN_1)        [right of=layer_norm_1,xshift=0.4cm] {GRU};
    \node[draw=black,rounded corners]     (RNN_2)        [right of=layer_norm_2,xshift=0.4cm] {GRU};
    \node[state]     (h12)    [right of=RNN_1,xshift=0.7cm]    {$\textbf{h}_1^{t+1}$};
    \node[state]     (h22)    [right of=RNN_2,xshift=0.7cm]    {$\textbf{h}_2^{t+1}$};

    \path (h11_label)   edge[dotted]                       (h11);
    \path (h21_label)   edge[dotted]                       (h21);
    \path (h11_label)   edge[short color dash,out=40,in=180]  (h10_embed); 
    \path (h21_label)   edge[short color dash,out=-40,in=180] (h20_embed); 
    
    \path (h11)         edge[out=-30,in=175,draw=blue!80]  (pool);
    \path (h11)         edge[out=-20,in=165,draw=green!80]  (pool);
    \path (h11)         edge[out=-10,in=155,draw=red!80]  (pool);
    \path (h21)         edge[out=30,in=185,draw=blue!180]  (pool);
    \path (h21)         edge[out=20,in=195,draw=red!180]  (pool);
    \path (h21)         edge[out=10,in=205,draw=green!180]  (pool);
    \path (pool)       edge[out=-10,in=210,draw=blue!80]  (layer_norm_1);
    \path (pool)       edge[out=0,in=200,draw=red!80]  (layer_norm_1);
    \path (pool)       edge[out=10,in=190,draw=green!80]  (layer_norm_1);
    \path (pool)       edge[out=-10,in=170,draw=blue!80]  (layer_norm_2);
    \path (pool)       edge[out=0,in=150,draw=red!80]  (layer_norm_2);
    \path (pool)       edge[out=10,in=160,draw=green!80]  (layer_norm_2);
    \path (layer_norm_1) edge[color dash]     (RNN_1);
    \path (layer_norm_2) edge[color dash]    (RNN_2);
    \path (h10_embed) edge[out=0,in=160,color dash] (RNN_1);
    \path (h20_embed) edge[out=0,in=200,color dash] (RNN_2);

    \path (RNN_1)         edge[color dash]    (h12);
    \path (RNN_2)         edge[color dash]    (h22);
  \end{tikzpicture}
  \caption{In each layer of the encoder after the initial BiGRU and linear transformation, hidden states corresponding to each atom are pooled across encodings of different SMILES strings for a common molecule, followed by layer norm and a GRU on each SMILES encoding independently.}
  \label{fig:encoder_block}
\end{figure}

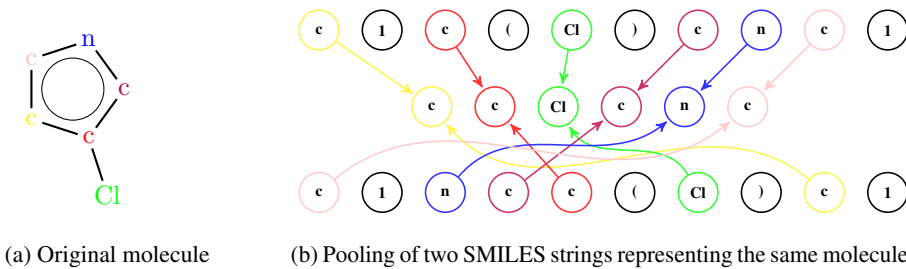
\begin{figure}[tbh]
  \centering
  \renewcommand * \printatom[1]{\ensuremath{\mathrm{#1}}}
  \setatomsep{8mm}
  
  \subfloat[Original molecule]{ 
    \adjustbox{valign=B,raise=\baselineskip}{
      \global\setbondstyle{thick,color=black}
      \chemfig{
        {\color{yellow} c}**5(-{\color{red} c}(-{\color{green}Cl})-{\color{purple} c}-{\color{blue} n}-{\color{pink} c}-)
      }
    }
    \hspace{0.8cm}
  }%
  \hspace{1cm}
  \subfloat[Pooling of two SMILES strings representing the same molecule]{
    \begin{tikzpicture}[->,>=stealth',shorten >=1pt,auto,node distance=1.4cm,semithick,font=\bfseries]
      \tikzstyle{every state}=[fill=white,draw=black,text=black,scale=0.6]
      \node[state,draw=yellow!80]  (c1)                       {c};
      \node[state]                 (12)     [right of=c1]     {1};
      \node[state,draw=red!80]     (c3)     [right of=12]     {c};
      \node[state]                 ((4)     [right of=c3]     {(};
      \node[state,draw=green!80]   (Cl5)    [right of=(4]     {Cl};
      \node[state]                 (6)      [right of=Cl5]    {)};
      \node[state,draw=purple!80]  (c7)     [right of=6]      {c};
      \node[state,draw=blue!80]    (n8)     [right of=c7]     {n};
      \node[state,draw=pink!80]    (c9)     [right of=n8]     {c};
      \node[state]                 (110)    [right of=c9]     {1};
      \node[state,draw=yellow!80]  (c1ag)   [xshift=2.5cm, yshift=-0.3cm,below of=c1] {c};
      \node[state,draw=red!80]     (c2ag)   [right of=c1ag]   {c};
      \node[state,draw=green!80]   (cl3_ag) [right of=c2ag]   {Cl};
      \node[state,draw=purple!80]  (c4ag)   [right of=cl3_ag] {c};
      \node[state,draw=blue!80]    (n5ag)   [right of=c4ag]   {n};
      \node[state,draw=pink!80]    (c6ag)   [right of=n5ag]   {c};
      \node[state,draw=pink!80]    (c1_2)   [yshift=-2.2cm,below of=c1] {c};
      \node[state]                 (12_2)   [right of=c1_2]    {1};
      \node[state,draw=blue!80]    (n3_2)   [right of=12_2]    {n};
      \node[state,draw=purple!80]  (c4_2)   [right of=n3_2]    {c};
      \node[state,draw=red!80]     (c5_2)   [right of=c4_2]    {c};
      \node[state]                 ((6_2)   [right of=c5_2]    {(};
      \node[state,draw=green!80]   (Cl7_2)  [right of=(6_2]    {Cl};
      \node[state]                 (8_2)    [right of=Cl7_2]   {)};
      \node[state,draw=yellow!80]  (c9_2)   [right of=8_2]     {c};
      \node[state]                 (110_2)  [right of=c9_2]    {1};
      \path (c1)     edge  [draw=yellow!80]                 (c1ag);
      \path (c9_2)   edge  [out=140,in=310,draw=yellow!80]  (c1ag); 
      \path (c3)     edge  [draw=red!80]                    (c2ag);
      \path (c5_2)   edge  [draw=red!80]                    (c2ag);
      \path (Cl5)    edge  [draw=green!80]                  (cl3_ag);
      \path (Cl7_2)  edge  [out=120,in=300,draw=green!80]   (cl3_ag); 
      \path (c7)     edge  [draw=purple!80]                 (c4ag);
      \path (c4_2)   edge  [draw=purple!80]                 (c4ag); 
      \path (n8)     edge  [draw=blue!80]                   (n5ag);
      \path (n3_2)   edge  [out=50,in=220,draw=blue!80]     (n5ag); 
      \path (c9)     edge  [draw=pink!80]                   (c6ag);
      \path (c1_2)   edge  [out=40,in=220,draw=pink!80]      (c6ag); %
    \end{tikzpicture}%
  }
  \caption{To pass information between distinct paths implicit in multiple SMILES representations of a molecule, the encoder pools the representation of each atom across multiple SMILES strings.}
  \label{fig:atom_aggregation}
\end{figure}

The characters of the multiple SMILES strings are linearly embedded, and each string is preprocessed by a BiGRU~\cite{cho2014learning}, followed by a linear transformation, to produce the initial hidden representation $\mathbf{h}_i^0$ for each SMILES string $i$.
The encoder then applies
a stack of modules, each of which pools between homologous atoms in the parallel representations, followed by layer norm, concatenation with the linearly embedded SMILES input, and a GRU applied to the parallel representations independently, as shown in Figures~\ref{fig:encoder_block} and~\ref{fig:atom_aggregation}. Each such parallel representation comprises a sequence of vectors, one for each character in the original SMILES string.

Multiple SMILES strings representing a single molecule need not have the same length, and syntactic characters indicating branching and ring closures rather than atoms and bonds do not generally match. However, the set of atoms is always consistent,
and a bijection can be defined between homologous atom characters. 
At the beginning of each encoder module (Figure~\ref{fig:encoder_block}), the parallel inputs corresponding to a single, common atom of the original molecule are pooled, as shown in Figure~\ref{fig:atom_aggregation}. This harmonized representation of the atom replaces the original in each of the input streams for the subsequent layer normalizations and GRUs, reversing the information flow of Figure~\ref{fig:atom_aggregation}. While we experimented with average and max pooling, we found element-wise sigmoid gating to be most effective~\cite{dauphin2017language, li2015gated, ryu2018deeply}: $a' = \frac{1}{k} \sum_k \left( a_k \odot \sigma \left( W \left[a_k, \frac{1}{k} \sum_k a_k \right] + b \right) \right)$, 
where $[x, y]$ is the concatenation of vectors $x$ and $y$, and the logistic function $\sigma(x) = \left(1 + e^{-x} \right)^{-1}$ is applied element-wise. 
The pooling effectively sums messages propagated from many adjacent nodes in the molecular graph, analogous to a graph convolution, but the GRUs efficiently transfer information through many edges in each layer, rather than just one.
The hidden representations associated with non-atom, syntactic input characters, such as parentheses and digits, are left unchanged by the pooling operation.

\begin{figure}[tbh]
  \center
  \begin{tikzpicture}[->,>=stealth',shorten >=1pt,auto,node distance=1.3cm,semithick]
    \tikzstyle{every state}=[fill=white,draw=black,text=black,scale=1.0,minimum size=0.1cm]
    \node[draw=black,rounded corners]   (GRU_1)                                        {$\text{GRU}_1$};
    \node[draw=black,rounded corners]   (GRU_2)     [below of=GRU_1]                   {$\text{GRU}_2$};
    \node[state,scale=0.8]              (before_GRU_1)  [left of=GRU_1,xshift=-0.5cm]                {$\textbf{h}_1^T$};
    \node[state,scale=0.8]              (before_GRU_2)  [left of=GRU_2,xshift=-0.5cm]                {$\textbf{h}_2^T$};
    \node[draw=black,rounded corners]   (max_pool)      [right of=GRU_1,yshift=-0.675cm]   {Pool};
    \node[draw=black,rounded corners]   (linear)        [right of=max_pool]                      {Lin};
    
    \node[state]                                         (z1)                    [right of=linear]                            {$\textbf{z}_1$};

    \node[draw=black,rounded corners]   (pool_atoms)       [above of=linear]                   {Pool atoms};
    
    \node[draw=black,rounded corners]   (q_NN_1)              [right of=z1]                                 {NN};
    \node[draw=black,rounded corners]   (attention_1)        [right of=q_NN_1]                         {Att};
    \node[state]                                         (z2)                    [right of=attention_1]                      {$\textbf{z}_2$};

    \node[draw=black,rounded corners]   (q_NN_2)              [right of=z2]                                 {NN};
    \node[draw=black,rounded corners]   (attention_2)        [right of=q_NN_2]                         {Att};
    \node[state]                                         (z3)                    [right of=attention_2]                      {$\textbf{z}_3$};

    \path (before_GRU_1) edge[short color dash]     (GRU_1);
    \path (before_GRU_2) edge[short color dash]    (GRU_2);
    \path (GRU_1) edge[out=-40, in=170]  (max_pool);
    \path (GRU_2) edge[out=40, in=190]    (max_pool);

    \path (GRU_1) edge[out=20, in=170,draw=green!80]     (pool_atoms);
    \path (GRU_1) edge[out=10, in=180,draw=red!80]     (pool_atoms);
    \path (GRU_1) edge[out=0, in=190,draw=blue!80]     (pool_atoms);

   \begin{pgfonlayer}{background}    
      \path (GRU_2) edge[out=75, in=200,draw=red!80]     (pool_atoms);
      \path (GRU_2) edge[out=65, in=210,draw=green!80]     (pool_atoms);
      \path (GRU_2) edge[out=55, in=220,draw=blue!80]     (pool_atoms);
    \end{pgfonlayer}

    \path (pool_atoms) edge[out=-20, in=150,draw=blue!80]     (attention_1);
    \path (pool_atoms) edge[out=-10, in=140,draw=red!80]     (attention_1);
    \path (pool_atoms) edge[out=0, in=130,draw=green!80] node[above,pos=0.97] {\color{black} k}    (attention_1);

    \path (pool_atoms) edge[out=-20, in=150,draw=blue!80]     (attention_2);
    \path (pool_atoms) edge[out=-10, in=140,draw=red!80]     (attention_2);
    \path (pool_atoms) edge[out=0, in=130,draw=green!80] node[above,pos=0.98] {\color{black} k}    (attention_2);

    \path (max_pool) edge (linear);
    \path (linear) edge node[below] {$\mu,\sigma$}   (z1);

    \path (z1) edge   (q_NN_1);
    \path (q_NN_1) edge node[below,xshift=-0.1cm] {q}  (attention_1);

    \path (attention_1) edge node[below] {$\mu,\sigma$}  (z2);
    \path (z2) edge (q_NN_2);
    \path (z1) edge[out=-25, in=-155] (q_NN_2);
    \path (q_NN_2) edge node[below,xshift=-0.1cm] {q} (attention_2);
    \path (attention_2) edge node[below] {$\mu,\sigma$} (z3);
  \end{tikzpicture}
  \caption{The approximating posterior is an autoregressive set of Gaussian distributions. The mean~($\mu$) and log-variance~($\log \sigma^2$) of the first subset of latent variables~$\textbf{z}_1$ is a linear transformation of the max-pooled final hidden state of GRUs fed the encoder outputs. Succeeding subsets~$\textbf{z}_i$ are produced via Bahdanau-style attention with the pooled atom outputs of the GRUs as keys~($k$), and the query~($q$) computed by a neural network on~$\textbf{z}_{<i}$.}
  \label{fig:approx_post}
\end{figure}
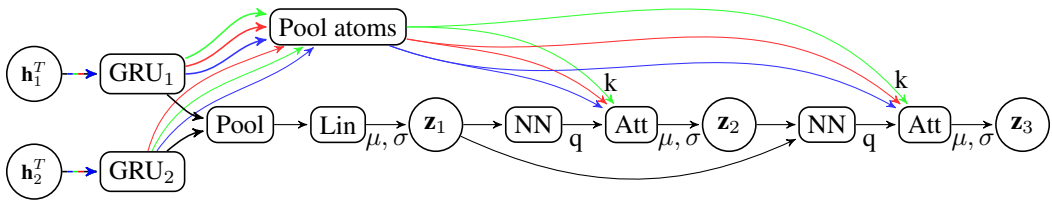

The approximating posterior distills the resulting variable-length encodings into a fixed-length hierarchy of autoregressive Gaussian distributions~~\cite{rolfe2016discrete}. 
The mean and log-variance of the first layer of the approximating posterior, $\mathbf{z}_1$, is parametrized by max-pooling the terminal hidden states of the final encoder GRUs, followed by batch renormalization~\cite{ioffe2017batch} and a linear transformation, as shown in Figure~\ref{fig:approx_post}. Succeeding hierarchical layers use Bahdanau-style attention~\cite{bahdanau2014neural} (reviewed in Appendix~\ref{sec:bahdanau_attention}) over the pooled final atom vectors, with the query vector defined by a one-hidden-layer network of rectified linear units (ReLUs) given the concatenation of the previous latent layers as input. 
This is analogous to the order-invariant encoding of set2set, but an output is produced at each step, and processing is not gated~\cite{vinyals2015order}. The attentional mechanism is also effectively available to property regressors that take the fixed-length latent representation as input, allowing them to aggregate contributions from across the molecule.
The output of the attentional mechanism is subject to batch renormalization and a linear transformation to compute the conditional mean and log-variance of the layer.
The prior has a similar autoregressive structure, but uses neural networks of ReLUs in place of Bahdanau-style attention, since it does not have access to the atom vectors.
For molecular optimization tasks, we usually scale up the term $\KL\left[ q(z|x) || p(z) \right]$ in the ELBO by the number of SMILES strings in the decoder,
analogous to multiple single-SMILES VAEs in parallel; we leave this KL term unscaled for property prediction.

The decoder is a single-layer LSTM, for which the initial cell state is computed from the latent representation by a neural network, and a linear transformation of the latent representation is concatenated onto each input.  
It is trained with teacher forcing to reconstruct a set of SMILES strings disjoint from those provided to the encoder, but representing the same molecule. Grammatical constraints~\cite{dai2018syntax, kusner2017grammar} can naturally be enforced within this LSTM by parsing the unfolding character sequence with a pushdown automaton, and constraining the final softmax of the LSTM output at each time step to grammatically valid symbols. This is detailed in Appendix~\ref{sec:SMILES_grammar}, although we leave the exploration of this technique to future work. 

Since the SMILES inputs to the encoder are different from the targets of the decoder, the decoder is effectively trained to assign equal probability to all SMILES strings of the encoded molecule. The latent representation must capture the molecule as a whole, rather than any particular SMILES input to the encoder. To accommodate this intentionally difficult reconstruction task, facilitate the construction of a bijection between latent space and molecules, and following prior work~\cite{kang2018conditional, winter2019learning}, we use a width-5 beam search decoder to map from the latent representation to the space of molecules at test-time. Further architectural details are presented in Appendix~\ref{sec:extended_model_architecture}.

\subsection{Latent space optimization}

Unlike many models that apply a sparse Gaussian process to fixed latent representations to predict molecular properties~\cite{dai2018syntax, jin2018junction, kusner2017grammar, samanta2018nevae}, the All SMILES VAE jointly trains property regressors with the generative model~\cite{liu2018constrained}.\footnote{\citet{gomez2018automatic} jointly train a regressor, but still optimize using a Gaussian process.}
We use linear regressors for the log octanol-water partition coefficient (logP) and molecular weight (MW), which have unbounded values; and logistic regressors for the quantitative estimate of drug-likeness (QED)~\cite{bickerton2012quantifying} and twelve binary measures of toxicity~\cite{huang2016tox21challenge, mayr2016deeptox}, which take values in~$\left[0, 1 \right]$.  We then perform gradient-based optimization of the property of interest with respect to the latent space, and decode the result to produce an optimized molecule.

Naively, we might either optimize the predicted property without constraints on the latent space, or find the maximum a posteriori (MAP) latent point for a conditional likelihood over the property that assigns greater probability to more desirable values.
However, the property regressors and decoder are only accurate within the domain in which they have been trained: the region assigned high probability mass by the prior. For a $n$-dimensional standard Gaussian prior, almost all probability mass lies in a practical support comprising a thin spherical shell of radius $\sqrt{n-1}$~\cite[][Gaussian Annulus Theorem]{foundations-of-data-science-2}.
With linear or logistic regressors, predicted property values increase monotonically in the direction of the weight vector, so unconstrained property maximization diverges from the origin of the latent space. Conversely, MAP optimization with a Gaussian prior is pulled towards the origin, where the density of the prior is greatest.
Both unconstrained and MAP optimization thus deviate from the practical support in each layer of the hierarchical prior, resulting in large prediction errors and poor optimization.

We can use the reparametrization trick~\cite{kingma2013auto, rezende2014stochastic} to map our autoregressive prior back to a standard Gaussian. The image of the thin spherical shell through this reparametrization still contains almost all of the probability mass. We therefore constrain optimization to the reparametrized $n-1$ dimensional sphere of radius $\sqrt{n-1}$ for each $n$-dimensional layer of the hierarchical prior by optimizing the angle directly.\footnote{This is generalizes the slerp interpolations of \citet{gomez2018automatic} to optimization.} 
Although the reparametrization from the standard Gaussian prior to our autoregressive prior is not volume preserving, this hierarchical radius constraint 
holds us to the center of the image of the thin spherical shell. The distance to which the image of the thin spherical shell extends away from the $n-1$ dimensional sphere at its center is a highly nonlinear function of the previous layers.

\section{Results} \label{sec:results}

We evaluate the All SMILES VAE on standard 250,000 and 310,000 element subsets~\cite{gomez2018automatic, kang2018conditional} of the ZINC database of small organic molecules~\cite{irwin2012zinc, sterling2015zinc}. We also evaluate on the Tox21 dataset~\cite{huang2016tox21challenge, mayr2016deeptox} in the DeepChem package~\cite{wu2018moleculenet}, comprising binarized binding affinities of 7831~compounds against 12~proteins. For further details, see Appendix~\ref{sec:datasets}. Additional experiments, including ablations of novel model components, are described in Appendix~\ref{sec:extended_results}.

The full power of continuous, gradient-based optimization can be brought bear on molecular properties given a bijection between molecules and contractible regions of a latent space, 
along with a regressor from the latent space to the property of interest that is differentiable almost everywhere. 
Such a bijection is challenging to confirm, since it is difficult to find the full latent space preimage of a molecule implicitly defined by a mapping from latent space to SMILES strings, such as our beam search decoder. As a necessary condition, we confirm that it is possible to map from the space of molecules to latent space and back again, and that random samples from the prior distribution in the latent space map to valid molecules. The former is required for injectivity, and the latter for surjectivity, of the mapping from molecules to contractible regions of the latent space.

Using the approximating posterior as the encoder, but always selecting the mean of each conditional Gaussian distribution (the maximum conditional a posteriori point), and a using beam search over the conditional likelihood as the decoder, 87.4\% $\pm$ 1\% of a held-out test set of ZINC250k (80/10/10 train/val/test split) is reconstructed accurately. With the same beam search decoder, 98.5\% $\pm$ 0.1\% of samples from the prior decode to valid SMILES strings. We expect that enforcing grammatical constraints in the decoder LSTM, as described in Appendix~\ref{sec:SMILES_grammar}, would further increase these rates. All molecules decoded from a set of 50,000 independent samples from the prior were unique, 99.958\% were novel relative to the training dataset,
and their average synthetic accessibility score~\cite{ertl2009estimation} was $2.97 \pm 0.01$, compared to $3.05$ in the ZINC250k dataset used for training.

\subsection{Property prediction}

Ultimately, we would like to optimize molecules for complicated physical properties, such as fluorescence quantum yield, solar cell efficiency, binding affinity to selected receptors, 
and low toxicity. 
Networks can only be trained to predict such physical properties if their true values are known on an appropriate training dataset.
While proxy properties can be accurately computed from first principles, properties like drug efficacy arise from highly nonlinear, poorly characterized processes, and can only be accurately determined through time-consuming and expensive experimental measurements. Since such experiments can only be performed on a tiny fraction of the $10^{60}$ drug-like molecules, we evaluate the ability of the All SMILES VAE to perform semi-supervised property prediction.

As Figure~\ref{fig:semi-supervised} and Table~\ref{tbl:semi_supervised_logp} in Appendix~\ref{sec:extended_results} demonstrate, we significantly improve the state-of-the-art in the semi-supervised prediction of simple molecular properties, including the log octanol-water partition coefficient~(logP), molecular weight~(MW), and quantitative estimate of drug-likeness~(QED)~\cite{bickerton2012quantifying}, against which many algorithms have been benchmarked. 
We achieve a similar improvement in fully supervised property prediction, as shown in Table~\ref{tbl:fully_supervised}, where we compare to extended connectivity fingerprints (ECFP)~\cite{rogers2010extended}, the character VAE (CVAE)~\cite{gomez2018automatic}, and graph convolutions~\cite{duvenaud2015convolutional}. Table~\ref{tbl:supervised_logp_with_pretrained_generative_model} in Appendix~\ref{sec:extended_results} documents an even larger improvement compared to models that use a sparse Gaussian process for property prediction.
We also surpass the state-of-the-art in toxicity prediction on the Tox21 dataset~\cite{huang2016tox21challenge, mayr2016deeptox}, as shown in Table~\ref{tbl:fully_supervised}, despite refraining from ensembling our model, or engineering features using expert chemistry knowledge, as in previous state-of-the-art methods~\cite{zaslavskiy2019toxicblend}. 

\begin{figure}[tbh]
  \center
  \subfloat[True logP over a 2D slice of latent space]{ \label{fig:heat_map}
    \includegraphics[height=4.7cm,trim={1.3cm 0.6cm 1.6cm 1.4cm},clip]{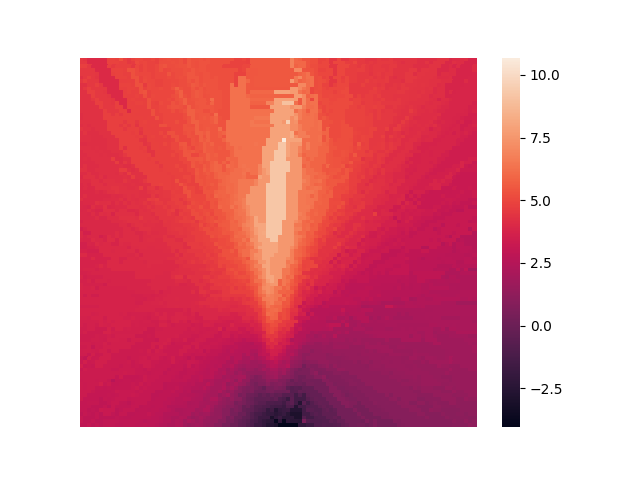}
  } \hfill
  \subfloat[Predicted and true logP over optimization]{ \label{fig:optimization_trajectory}
    \includegraphics[height=4.7cm,trim={1cm 0.2cm 1cm 1.2cm},clip]{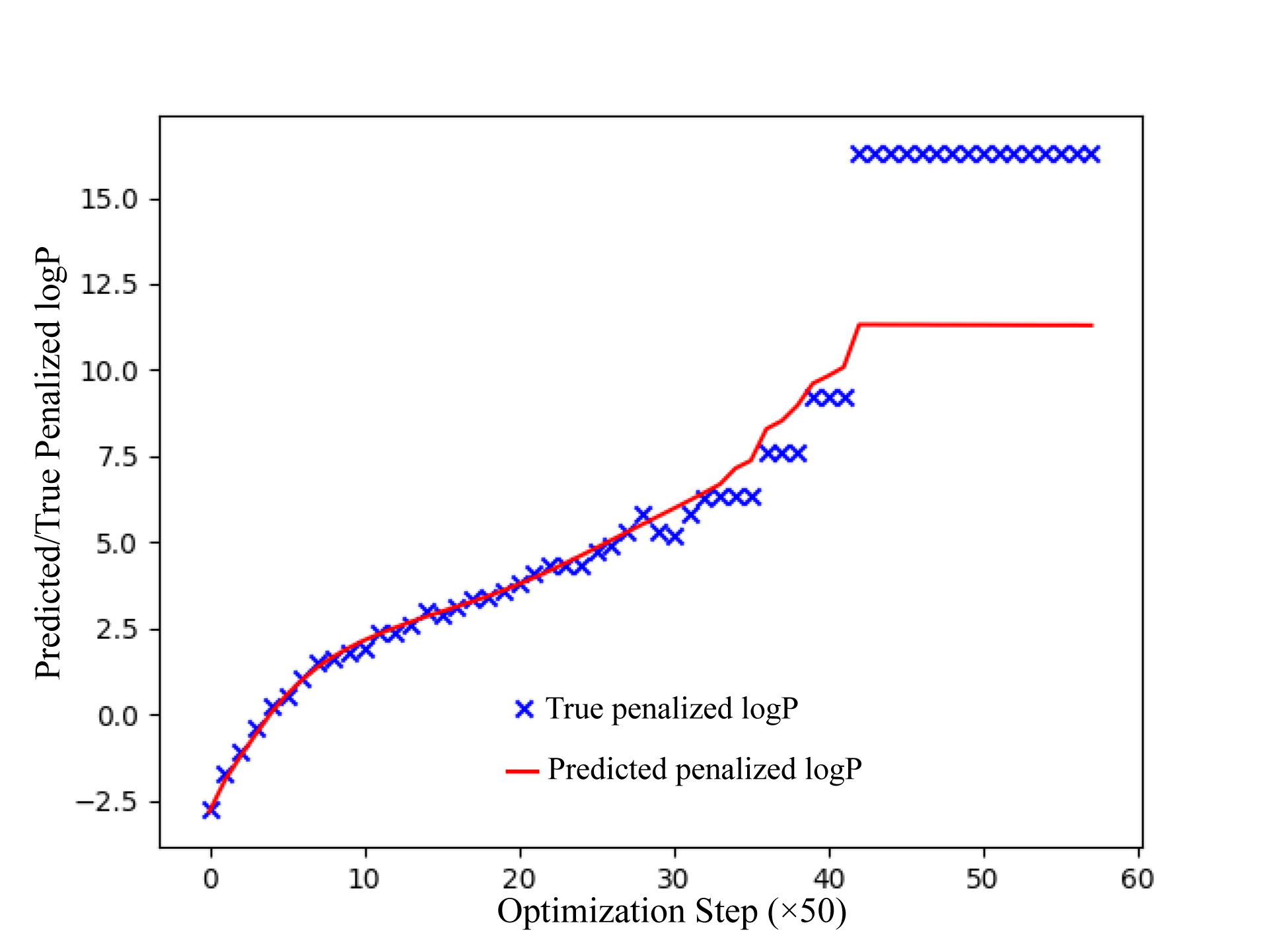}
  }
  \caption{Dense decodings of true logP along a local 2D sheet in latent space, with the y axis aligned with the regressor (a), and predicted and true penalized logP across steps of optimization (b).}
  \label{fig:smoothness}
\end{figure}

Accurate property prediction only facilitates effective optimization if the true property value is smooth with respect to the latent space. In Figure~\ref{fig:heat_map}, we plot the true (not predicted) logP over a densely sampled 2D slice of the latent space, where the y axis is aligned with the logP linear regressor.

\pgfplotsset{every tick label/.append style={font=\tiny}}

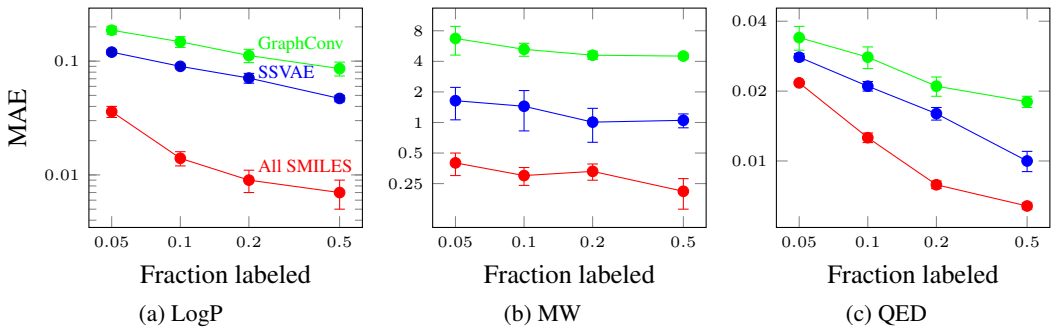
\begin{figure}[tbh]
  \centering
  \subfloat[LogP]{%
    \begin{tikzpicture}%
      \begin{loglogaxis}[
        height=4.5cm,
          xlabel={Fraction labeled},
          ylabel={MAE},
          xlabel near ticks,
          ylabel near ticks,
          xtick={0.05, 0.1, 0.2, 0.5},
          log ticks with fixed point]
        \addplot+[error bars/.cd,y dir=both,y explicit] coordinates {
          (0.05, 0.120) +- (0.006,0.006)
          (0.10, 0.090) +- (0.004,0.004)
          (0.20, 0.071) +- (0.007,0.007)
          (0.50, 0.047) +- (0.003,0.003)
        };
        \addplot+[color=green,mark=*,mark options={green},error bars/.cd,y dir=both,y explicit] coordinates { 
          (0.05, 0.187) +- (0.015,0.015)
          (0.10, 0.148) +- (0.016,0.016)
          (0.20, 0.112) +- (0.015,0.015)
          (0.50, 0.086) +- (0.012,0.012)
        };
        \addplot[color=red,mark=*,mark options={red},error bars/.cd,y dir=both,y explicit] coordinates {
          (0.05, 0.036) +- (0.004,0.004)
          (0.10, 0.014) +- (0.002,0.002)
          (0.20, 0.009) +- (0.002,0.002)
          (0.50, 0.007) +- (0.002,0.002)
        };
        \node[right, blue] at (axis cs:0.2, 0.08) {\scriptsize SSVAE};%
        \node[right, green] at (axis cs:0.2, 0.14) {\scriptsize GraphConv};%
        \node[right, red] at (axis cs:0.2, 0.012) {\scriptsize All SMILES};%
      \end{loglogaxis}%
    \end{tikzpicture}%
  }
  \hfill
  \subfloat[MW]{%
    \begin{tikzpicture}%
      \begin{loglogaxis}[
        height=4.5cm,
          xlabel={Fraction labeled},
          xlabel near ticks,
          xtick={0.05, 0.1, 0.2, 0.5},
          ytick={8.0, 4.0, 2.0, 1.0, 0.5, 0.25},
          log ticks with fixed point]
        \addplot+[color=blue,mark=*,mark options={blue},error bars/.cd,y dir=both,y explicit] coordinates {
          (0.05, 1.639) +- (0.577,0.577)
          (0.10, 1.444) +- (0.618,0.618)
          (0.20, 1.008) +- (0.370,0.370)
          (0.50, 1.05) +- (0.164,0.164)
        };
        \addplot+[color=green,mark=*,mark options={green},error bars/.cd,y dir=both,y explicit] coordinates { 
          (0.05, 6.723) +- (2.116,2.116)
          (0.10, 5.255) +- (0.767,0.767)
          (0.20, 4.597) +- (0.419,0.419)
          (0.50, 4.506) +- (0.279,0.279)
        };
        \addplot[color=red,mark=*,mark options={red},error bars/.cd,y dir=both,y explicit] coordinates {
          (0.05, 0.4) +- (0.1,0.1)
          (0.10, 0.30) +- (0.06,0.06)
          (0.20, 0.33) +- (0.06,0.06)
          (0.50, 0.21) +- (0.07,0.07)
        };
      \end{loglogaxis}%
    \end{tikzpicture}%
  }
  \hfill
  \subfloat[QED]{%
    \begin{tikzpicture}%
      \begin{loglogaxis}[
        height=4.5cm,
          xlabel={Fraction labeled},
          xlabel near ticks,
          xtick={0.05, 0.1, 0.2, 0.5},
          ytick={0.04, 0.02, 0.01, 0.005},
          log ticks with fixed point]
        \addplot+[color=blue,mark=*,mark options={blue},error bars/.cd,y dir=both,y explicit] coordinates {
          (0.05, 0.028) +- (0.001,0.001)
          (0.10, 0.021) +- (0.001,0.001)
          (0.20, 0.016) +- (0.001,0.001)
          (0.50, 0.01) +- (0.001,0.001)
        };
        \addplot+[color=green,mark=*,mark options={green},error bars/.cd,y dir=both,y explicit] coordinates { 
          (0.05, 0.034) +- (0.004,0.004)
          (0.10, 0.028) +- (0.003,0.003)
          (0.20, 0.021) +- (0.002,0.002)
          (0.50, 0.018) +- (0.001,0.001)
        };
        \addplot[color=red,mark=*,mark options={red},error bars/.cd,y dir=both,y explicit] coordinates {
          (0.05, 0.0217) +- (0.0003,0.0003)
          (0.10, 0.0126) +- (0.0006,0.0006)
          (0.20, 0.0079) +- (0.0003,0.0003)
          (0.50, 0.0064) +- (0.0002,0.0002)
        };
      \end{loglogaxis}%
    \end{tikzpicture}%
  }
  \caption{Semi-supervised mean absolute error (MAE) $\pm$ the standard deviation across ten replicates for the log octanol-water partition coefficient (a), molecular weight (b), and the quantitative estimate drug-likeness~\cite{bickerton2012quantifying} (c) on the ZINC310k dataset. Plots are log-log; the All SMILES MAE is a fraction of that of the SSVAE~\cite{kang2018conditional} and graph convolutions~\cite{kearnes2016molecular}. Semi-supervised VAE (SSVAE) and graph convolution results are those reported by \citet{kang2018conditional}.}%
  \label{fig:semi-supervised}%
\end{figure}

\begin{table}[tbh]
  \caption{Fully supervised regression on ZINC250k (a), evaluated using the mean absolute error; and Tox21 (b), evaluated with the area under the receiver operating characteristic curve (AUC-ROC), averaged over all 12 toxicity types. 
    Aside from All SMILES, results in (a) are those reported in~\cite{gomez2018automatic}. 
} 
  \label{tbl:fully_supervised}
  \centering
  \subfloat[ZINC250k]{
      \begin{sc}
        \begin{tabular}{lll}
          \toprule
          Model & MAE logP & MAE QED\\
          \midrule
          ECFP~\cite{rogers2010extended} & 0.38  & 0.045 \\ 
          CVAE~\cite{gomez2018automatic} & 0.15  & 0.054 \\
          CVAE enc~\cite{gomez2018automatic} & 0.13  & 0.037 \\
          GraphConv~\cite{duvenaud2015convolutional}  & 0.05 & 0.017 \\
          {\bf All SMILES}               & {\bf 0.005 {\tiny $\pm$ 0.0006}} & {\bf 0.0052 {\tiny $\pm$ 0.0001}} \\
          \bottomrule
        \end{tabular}
      \end{sc}
  }
  \subfloat[Tox21]{
      \begin{sc}
        \begin{tabular}{ll}
          \toprule
          Model & AUC-ROC    \\
          \midrule
          GraphConv~\cite{wu2018moleculenet} & 0.829 {\tiny $\pm$ 0.006} \\
          Li, Cai, \& He~\cite{li2017learning} & 0.854 \\
          PotentialNet~\cite{feinberg2018potentialnet} & 0.857 {\tiny $\pm$ 0.006} \\
          ToxicBlend~\cite{zaslavskiy2019toxicblend} & 0.862 \\
          {\bf All SMILES} &  {\bf 0.871} \\
          \bottomrule
        \end{tabular}
      \end{sc}
  }
\end{table}

\subsection{Molecular optimization} \label{sec:molecular_optimization}

\begin{figure}[tbh]
  \centering
  \subfloat[Molecules with the top three penalized logP values]{
    \begin{tabular}{ccc}
      \includegraphics[width=1.83cm,trim={1.2cm 2.5cm 1.2cm 2.4cm},clip]{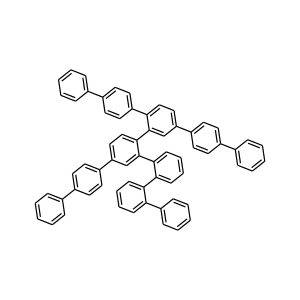} &
      \includegraphics[width=1.83cm,trim={1.2cm 3.0cm 1.2cm 3.5cm},clip]{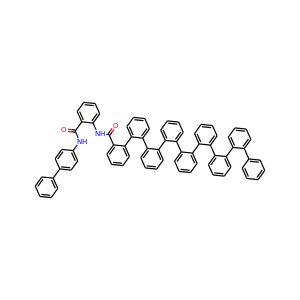} &
      \includegraphics[width=1.83cm,trim={1.2cm 3.0cm 1.2cm 3.6cm},clip]{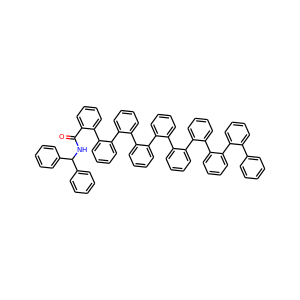}
    \end{tabular}
  }
  \hfill
  \subfloat[Molecules with the top three QED values]{
    \begin{tabular}{ccc}
      \includegraphics[width=1.83cm,trim={1.2cm 3.0cm 1.2cm 3.5cm},clip]{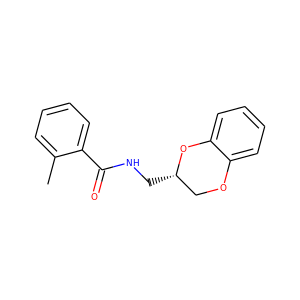} &
      \includegraphics[width=1.83cm,trim={1.2cm 3.0cm 1.2cm 3.6cm},clip]{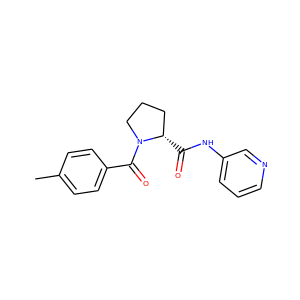} &
      \includegraphics[width=1.83cm,trim={1.2cm 2.9cm 1.2cm 2.9cm},clip]{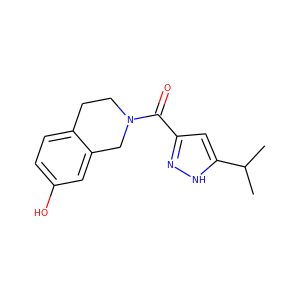}
    \end{tabular}
  }
  \caption{Molecules produced by gradient-based optimization 
    in the All SMILES VAE.}
  \label{fig:top_molecules}
\end{figure}

We maximize the output of our linear and logistic property regressors, plus a log-prior regularizer, with respect to the latent space, subject to a hierarchical radius constraint. After optimizing in the latent space with ADAM, we project back to a SMILES representation of a molecule with the decoder. Following prior work, we optimize QED and logP penalized by the synthetic accessibility score and the number of large rings~\cite{dai2018syntax, jin2018junction, kusner2017grammar, samanta2018nevae, you2018graph, zhou2018optimization}. Figure~\ref{fig:optimization_trajectory} depicts the predicted and true logP over an optimization trajectory, while Table~\ref{table:optimization_results} compares the top three values found amongst 100 such trajectories to the previous state-of-the-art.\footnote{Recently, \citet{winter2019efficient} reported molecules with penalized logP as large as 26.1, but train on an enormous, non-standard dataset of 72 million compounds aggregated from the ZINC15 and PubChem databases.} 
The molecules realizing these property values are shown in Figure~\ref{fig:top_molecules}. 
Leaving the KL term in the ELBO unscaled by the number of SMILES strings in the decoder reduces the regularization of the latent space embeddings, allowing latent space optimization to search a wider space of molecules that are less similar to the training set, as shown in Figure~\ref{fig:top_penlogp_lowkl} of Appendix~\ref{sec:extended_molecular_optimization}. 
Unlike reinforcement learning methods that progressively evaluate the properties of novel molecules generated during training, including Graph Convolutional Policy Networks (GCPN) and Molecule Deep Q-Networks (MolDQN)~\cite{you2018graph, zhou2018optimization}, the All SMILES VAE only requires a fixed training dataset. This is critical when optimizing properties, including pharmacological efficacy, toxicity, and OLED efficiency, for which accurate \emph{in silico} approximations do not exist.

\begin{table}[tbh]
  \centering
  \caption{Properties of the top three optimized molecules trained on ZINC250k.}
  \label{table:optimization_results}
  \subfloat {
      \begin{sc}
        \begin{tabular}{ll}
          \toprule
          Model & Penalized logP     \\
          \midrule
          JT-VAE~\cite{jin2018junction} & 5.30, 4.93, 4.49 \\
          GCPN~\cite{you2018graph} & 7.98, 7.85, 7.80 \\
          MolDQN~\cite{zhou2018optimization} & 11.84, 11.84, 11.82 \\
          All SMILES & 16.42, 16.32, 16.21 \\
          {\bf All SMILES {\tiny (KL unscaled)}} & {\bf 42.46, 42.42, 41.54} \\
          \bottomrule
        \end{tabular}
      \end{sc}
  }
  \hfill
  \subfloat{ 
      \begin{sc}
        \begin{tabular}{ll}
          \toprule
          Model & QED     \\
          \midrule
          JT-VAE~\cite{jin2018junction} & 0.925, 0.911, 0.910 \\ 
          CGVAE~\cite{liu2018constrained} & 0.938, 0.931, 0.880 \\
          GCPN~\cite{you2018graph} & 0.948, 0.947, 0.946 \\
          {\bf MolDQN}~\cite{zhou2018optimization} & {\bf 0.948, 0.948, 0.948} \\
          {\bf All SMILES} & {\bf 0.948, 0.948, 0.948} \\
          \bottomrule
        \end{tabular}
      \end{sc}
  }
\end{table}

\section{Conclusion}

For each molecule, the All SMILES encoder uses stacked, pooled RNNs on multiple SMILES strings to efficiently pass information throughout the molecular graph. The decoder targets a disjoint set of SMILES strings of the same molecule, forcing the latent space to develop a consistent representation for each molecule. Attentional mechanisms in the approximating posterior summarize spatially diffuse features into a fixed-length, non-factorial approximating posterior, and construct a latent representation on which linear regressors achieve state-of-the-art semi- and fully-supervised property prediction. Gradient-based optimization of these regressor outputs with respect to the latent representation, constrained to a subspace near almost all probability in the prior, produces state-of-the-art optimized molecules when coupled with a simple RNN decoder. 

\section*{Acknowledgements}

We thank Alla Pryyma for help in preparing Figures~\ref{fig:SMILES_def}, \ref{fig:qed_optimization_trajectory}, and~\ref{fig:optimization_trajectory}. Mani Ranjbar provided invaluable, unstinting assistance with the compute cluster. Juan Carrasquilla, Colin Collins, Faraz Hach, and William G. Macready contributed helpful discussions.

\bibliography{all_smiles_nips_arxiv}

\begin{thebibliography}{72}
\providecommand{\natexlab}[1]{#1}
\providecommand{\url}[1]{\texttt{#1}}
\expandafter\ifx\csname urlstyle\endcsname\relax
  \providecommand{\doi}[1]{doi: #1}\else
  \providecommand{\doi}{doi: \begingroup \urlstyle{rm}\Url}\fi

\bibitem[Appleyard et~al.(2016)Appleyard, Kocisky, and
  Blunsom]{appleyard2016optimizing}
Appleyard, J., Kocisky, T., and Blunsom, P.
\newblock Optimizing performance of recurrent neural networks on gpus.
\newblock \emph{arXiv preprint arXiv:1604.01946}, 2016.

\bibitem[Aumentado-Armstrong(2018)]{aumentado2018latent}
Aumentado-Armstrong, T.
\newblock Latent molecular optimization for targeted therapeutic design.
\newblock \emph{arXiv preprint arXiv:1809.02032}, 2018.

\bibitem[Bahdanau et~al.(2014)Bahdanau, Cho, and Bengio]{bahdanau2014neural}
Bahdanau, D., Cho, K., and Bengio, Y.
\newblock Neural machine translation by jointly learning to align and
  translate.
\newblock \emph{arXiv preprint arXiv:1409.0473}, 2014.

\bibitem[Bickerton et~al.(2012)Bickerton, Paolini, Besnard, Muresan, and
  Hopkins]{bickerton2012quantifying}
Bickerton, G.~R., Paolini, G.~V., Besnard, J., Muresan, S., and Hopkins, A.~L.
\newblock Quantifying the chemical beauty of drugs.
\newblock \emph{Nature chemistry}, 4\penalty0 (2):\penalty0 90, 2012.

\bibitem[Bjerrum \& Sattarov(2018)Bjerrum and Sattarov]{bjerrum2018improving}
Bjerrum, E. and Sattarov, B.
\newblock Improving chemical autoencoder latent space and molecular de novo
  generation diversity with heteroencoders.
\newblock \emph{Biomolecules}, 8\penalty0 (4):\penalty0 131, 2018.

\bibitem[Blum et~al.(2017)Blum, Hopcroft, and
  Kannan]{foundations-of-data-science-2}
Blum, A., Hopcroft, J., and Kannan, R.
\newblock \emph{Foundations of Data Science}.
\newblock June 2017.
\newblock URL
  \url{https://www.microsoft.com/en-us/research/publication/foundations-of-data-science-2/}.

\bibitem[Cho et~al.(2014)Cho, Van~Merri{\"e}nboer, Gulcehre, Bahdanau,
  Bougares, Schwenk, and Bengio]{cho2014learning}
Cho, K., Van~Merri{\"e}nboer, B., Gulcehre, C., Bahdanau, D., Bougares, F.,
  Schwenk, H., and Bengio, Y.
\newblock Learning phrase representations using rnn encoder-decoder for
  statistical machine translation.
\newblock \emph{arXiv preprint arXiv:1406.1078}, 2014.

\bibitem[Clayden et~al.(2001)Clayden, Greeves, Warren, and
  Wothers]{clayden2001organic}
Clayden, J., Greeves, N., Warren, S., and Wothers, P.
\newblock \emph{Organic chemistry}.
\newblock Oxford University Press, 2001.

\bibitem[Dai et~al.(2018)Dai, Tian, Dai, Skiena, and Song]{dai2018syntax}
Dai, H., Tian, Y., Dai, B., Skiena, S., and Song, L.
\newblock Syntax-directed variational autoencoder for structured data.
\newblock \emph{arXiv preprint arXiv:1802.08786}, 2018.

\bibitem[Dauphin et~al.(2017)Dauphin, Fan, Auli, and
  Grangier]{dauphin2017language}
Dauphin, Y.~N., Fan, A., Auli, M., and Grangier, D.
\newblock Language modeling with gated convolutional networks.
\newblock In \emph{Proceedings of the 34th International Conference on Machine
  Learning-Volume 70}, pp.\  933--941. JMLR. org, 2017.

\bibitem[De~Cao \& Kipf(2018)De~Cao and Kipf]{de2018molgan}
De~Cao, N. and Kipf, T.
\newblock Molgan: An implicit generative model for small molecular graphs.
\newblock \emph{arXiv preprint arXiv:1805.11973}, 2018.

\bibitem[Duvenaud et~al.(2015)Duvenaud, Maclaurin, Iparraguirre, Bombarell,
  Hirzel, Aspuru-Guzik, and Adams]{duvenaud2015convolutional}
Duvenaud, D.~K., Maclaurin, D., Iparraguirre, J., Bombarell, R., Hirzel, T.,
  Aspuru-Guzik, A., and Adams, R.~P.
\newblock Convolutional networks on graphs for learning molecular fingerprints.
\newblock In \emph{Advances in neural information processing systems}, pp.\
  2224--2232, 2015.

\bibitem[Ertl \& Schuffenhauer(2009)Ertl and Schuffenhauer]{ertl2009estimation}
Ertl, P. and Schuffenhauer, A.
\newblock Estimation of synthetic accessibility score of drug-like molecules
  based on molecular complexity and fragment contributions.
\newblock \emph{Journal of cheminformatics}, 1\penalty0 (1):\penalty0 8, 2009.

\bibitem[Feinberg et~al.(2018)Feinberg, Sur, Wu, Husic, Mai, Li, Sun, Yang,
  Ramsundar, and Pande]{feinberg2018potentialnet}
Feinberg, E.~N., Sur, D., Wu, Z., Husic, B.~E., Mai, H., Li, Y., Sun, S., Yang,
  J., Ramsundar, B., and Pande, V.~S.
\newblock Potentialnet for molecular property prediction.
\newblock \emph{ACS central science}, 4\penalty0 (11):\penalty0 1520--1530,
  2018.

\bibitem[Gilmer et~al.(2017)Gilmer, Schoenholz, Riley, Vinyals, and
  Dahl]{gilmer2017neural}
Gilmer, J., Schoenholz, S.~S., Riley, P.~F., Vinyals, O., and Dahl, G.~E.
\newblock Neural message passing for quantum chemistry.
\newblock In \emph{Proceedings of the 34th International Conference on Machine
  Learning-Volume 70}, pp.\  1263--1272. JMLR. org, 2017.

\bibitem[G{\'o}mez-Bombarelli et~al.(2018)G{\'o}mez-Bombarelli, Wei, Duvenaud,
  Hern{\'a}ndez-Lobato, S{\'a}nchez-Lengeling, Sheberla, Aguilera-Iparraguirre,
  Hirzel, Adams, and Aspuru-Guzik]{gomez2018automatic}
G{\'o}mez-Bombarelli, R., Wei, J.~N., Duvenaud, D., Hern{\'a}ndez-Lobato,
  J.~M., S{\'a}nchez-Lengeling, B., Sheberla, D., Aguilera-Iparraguirre, J.,
  Hirzel, T.~D., Adams, R.~P., and Aspuru-Guzik, A.
\newblock Automatic chemical design using a data-driven continuous
  representation of molecules.
\newblock \emph{ACS central science}, 4\penalty0 (2):\penalty0 268--276, 2018.

\bibitem[Guimaraes et~al.(2017)Guimaraes, Sanchez-Lengeling, Outeiral, Farias,
  and Aspuru-Guzik]{guimaraes2017objective}
Guimaraes, G.~L., Sanchez-Lengeling, B., Outeiral, C., Farias, P. L.~C., and
  Aspuru-Guzik, A.
\newblock Objective-reinforced generative adversarial networks (organ) for
  sequence generation models.
\newblock \emph{arXiv preprint arXiv:1705.10843}, 2017.

\bibitem[Gupta et~al.(2018)Gupta, M{\"u}ller, Huisman, Fuchs, Schneider, and
  Schneider]{gupta2018generative}
Gupta, A., M{\"u}ller, A.~T., Huisman, B.~J., Fuchs, J.~A., Schneider, P., and
  Schneider, G.
\newblock Generative recurrent networks for de novo drug design.
\newblock \emph{Molecular informatics}, 37\penalty0 (1-2):\penalty0 1700111,
  2018.

\bibitem[Hammond et~al.(2011)Hammond, Vandergheynst, and
  Gribonval]{hammond2011wavelets}
Hammond, D.~K., Vandergheynst, P., and Gribonval, R.
\newblock Wavelets on graphs via spectral graph theory.
\newblock \emph{Applied and Computational Harmonic Analysis}, 30\penalty0
  (2):\penalty0 129--150, 2011.

\bibitem[He et~al.(2016)He, Zhang, Ren, and Sun]{he2016deep}
He, K., Zhang, X., Ren, S., and Sun, J.
\newblock Deep residual learning for image recognition.
\newblock In \emph{Proceedings of the IEEE conference on computer vision and
  pattern recognition}, pp.\  770--778, 2016.

\bibitem[Hochreiter \& Schmidhuber(1997)Hochreiter and
  Schmidhuber]{hochreiter1997long}
Hochreiter, S. and Schmidhuber, J.
\newblock Long short-term memory.
\newblock \emph{Neural computation}, 9\penalty0 (8):\penalty0 1735--1780, 1997.

\bibitem[Huang et~al.(2016)Huang, Xia, Nguyen, Zhao, Sakamuru, Zhao, Shahane,
  Rossoshek, and Simeonov]{huang2016tox21challenge}
Huang, R., Xia, M., Nguyen, D.-T., Zhao, T., Sakamuru, S., Zhao, J., Shahane,
  S.~A., Rossoshek, A., and Simeonov, A.
\newblock Tox21challenge to build predictive models of nuclear receptor and
  stress response pathways as mediated by exposure to environmental chemicals
  and drugs.
\newblock \emph{Frontiers in Environmental Science}, 3:\penalty0 85, 2016.

\bibitem[Im et~al.(2017)Im, Kim, Cho, Seo, Yook, and Lee]{im2017molecular}
Im, Y., Kim, M., Cho, Y.~J., Seo, J.-A., Yook, K.~S., and Lee, J.~Y.
\newblock Molecular design strategy of organic thermally activated delayed
  fluorescence emitters.
\newblock \emph{Chemistry of Materials}, 29\penalty0 (5):\penalty0 1946--1963,
  2017.

\bibitem[Ioffe(2017)]{ioffe2017batch}
Ioffe, S.
\newblock Batch renormalization: Towards reducing minibatch dependence in
  batch-normalized models.
\newblock In \emph{Advances in neural information processing systems}, pp.\
  1945--1953, 2017.

\bibitem[Irwin et~al.(2012)Irwin, Sterling, Mysinger, Bolstad, and
  Coleman]{irwin2012zinc}
Irwin, J.~J., Sterling, T., Mysinger, M.~M., Bolstad, E.~S., and Coleman, R.~G.
\newblock Zinc: a free tool to discover chemistry for biology.
\newblock \emph{Journal of chemical information and modeling}, 52\penalty0
  (7):\penalty0 1757--1768, 2012.

\bibitem[Jaques et~al.(2017)Jaques, Gu, Bahdanau, Hern{\'a}ndez-Lobato, Turner,
  and Eck]{jaques2017sequence}
Jaques, N., Gu, S., Bahdanau, D., Hern{\'a}ndez-Lobato, J.~M., Turner, R.~E.,
  and Eck, D.
\newblock Sequence tutor: Conservative fine-tuning of sequence generation
  models with kl-control.
\newblock In \emph{Proceedings of the 34th International Conference on Machine
  Learning}, pp.\  1645--1654. JMLR.org, 2017.

\bibitem[Jin et~al.(2018)Jin, Barzilay, and Jaakkola]{jin2018junction}
Jin, W., Barzilay, R., and Jaakkola, T.
\newblock Junction tree variational autoencoder for molecular graph generation.
\newblock \emph{arXiv preprint arXiv:1802.04364}, 2018.

\bibitem[Kang \& Cho(2018)Kang and Cho]{kang2018conditional}
Kang, S. and Cho, K.
\newblock Conditional molecular design with deep generative models.
\newblock \emph{arXiv preprint arXiv:1805.00108}, 2018.

\bibitem[Kearnes et~al.(2016)Kearnes, McCloskey, Berndl, Pande, and
  Riley]{kearnes2016molecular}
Kearnes, S., McCloskey, K., Berndl, M., Pande, V., and Riley, P.
\newblock Molecular graph convolutions: moving beyond fingerprints.
\newblock \emph{Journal of computer-aided molecular design}, 30\penalty0
  (8):\penalty0 595--608, 2016.

\bibitem[Kim et~al.(2018)Kim, Chen, Cheng, Gindulyte, He, He, Li, Shoemaker,
  Thiessen, Yu, et~al.]{kim2018pubchem}
Kim, S., Chen, J., Cheng, T., Gindulyte, A., He, J., He, S., Li, Q., Shoemaker,
  B.~A., Thiessen, P.~A., Yu, B., et~al.
\newblock Pubchem 2019 update: improved access to chemical data.
\newblock \emph{Nucleic acids research}, 47\penalty0 (D1):\penalty0
  D1102--D1109, 2018.

\bibitem[Kingma \& Welling(2013)Kingma and Welling]{kingma2013auto}
Kingma, D.~P. and Welling, M.
\newblock Auto-encoding variational bayes.
\newblock \emph{arXiv preprint arXiv:1312.6114}, 2013.

\bibitem[Kipf \& Welling(2016{\natexlab{a}})Kipf and Welling]{kipf2016semi}
Kipf, T.~N. and Welling, M.
\newblock Semi-supervised classification with graph convolutional networks.
\newblock \emph{arXiv preprint arXiv:1609.02907}, 2016{\natexlab{a}}.

\bibitem[Kipf \& Welling(2016{\natexlab{b}})Kipf and
  Welling]{kipf2016variational}
Kipf, T.~N. and Welling, M.
\newblock Variational graph auto-encoders.
\newblock \emph{arXiv preprint arXiv:1611.07308}, 2016{\natexlab{b}}.

\bibitem[Krizhevsky et~al.(2012)Krizhevsky, Sutskever, and
  Hinton]{krizhevsky2012imagenet}
Krizhevsky, A., Sutskever, I., and Hinton, G.~E.
\newblock Imagenet classification with deep convolutional neural networks.
\newblock In \emph{Advances in neural information processing systems}, pp.\
  1097--1105, 2012.

\bibitem[Kusner et~al.(2017)Kusner, Paige, and
  Hern{\'a}ndez-Lobato]{kusner2017grammar}
Kusner, M.~J., Paige, B., and Hern{\'a}ndez-Lobato, J.~M.
\newblock Grammar variational autoencoder.
\newblock \emph{arXiv preprint arXiv:1703.01925}, 2017.

\bibitem[Landrum et~al.(2006)]{landrum2006rdkit}
Landrum, G. et~al.
\newblock Rdkit: Open-source cheminformatics, 2006.

\bibitem[LeCun et~al.(1990)LeCun, Boser, Denker, Henderson, Howard, Hubbard,
  and Jackel]{lecun1990handwritten}
LeCun, Y., Boser, B.~E., Denker, J.~S., Henderson, D., Howard, R.~E., Hubbard,
  W.~E., and Jackel, L.~D.
\newblock Handwritten digit recognition with a back-propagation network.
\newblock In \emph{Advances in neural information processing systems}, pp.\
  396--404, 1990.

\bibitem[Li et~al.(2017)Li, Cai, and He]{li2017learning}
Li, J., Cai, D., and He, X.
\newblock Learning graph-level representation for drug discovery.
\newblock \emph{arXiv preprint arXiv:1709.03741}, 2017.

\bibitem[Li et~al.(2015)Li, Tarlow, Brockschmidt, and Zemel]{li2015gated}
Li, Y., Tarlow, D., Brockschmidt, M., and Zemel, R.
\newblock Gated graph sequence neural networks.
\newblock \emph{arXiv preprint arXiv:1511.05493}, 2015.

\bibitem[Li et~al.(2018)Li, Vinyals, Dyer, Pascanu, and
  Battaglia]{li2018learning}
Li, Y., Vinyals, O., Dyer, C., Pascanu, R., and Battaglia, P.
\newblock Learning deep generative models of graphs.
\newblock \emph{arXiv preprint arXiv:1803.03324}, 2018.

\bibitem[Lim et~al.(2018)Lim, Ryu, Kim, and Kim]{lim2018molecular}
Lim, J., Ryu, S., Kim, J.~W., and Kim, W.~Y.
\newblock Molecular generative model based on conditional variational
  autoencoder for de novo molecular design.
\newblock \emph{Journal of cheminformatics}, 10\penalty0 (1):\penalty0 31,
  2018.

\bibitem[Liu et~al.(2018)Liu, Allamanis, Brockschmidt, and
  Gaunt]{liu2018constrained}
Liu, Q., Allamanis, M., Brockschmidt, M., and Gaunt, A.~L.
\newblock Constrained graph variational autoencoders for molecule design.
\newblock \emph{arXiv preprint arXiv:1805.09076}, 2018.

\bibitem[Lusci et~al.(2013)Lusci, Pollastri, and Baldi]{lusci2013deep}
Lusci, A., Pollastri, G., and Baldi, P.
\newblock Deep architectures and deep learning in chemoinformatics: the
  prediction of aqueous solubility for drug-like molecules.
\newblock \emph{Journal of chemical information and modeling}, 53\penalty0
  (7):\penalty0 1563--1575, 2013.

\bibitem[Ma et~al.(2018)Ma, Chen, and Xiao]{ma2018constrained}
Ma, T., Chen, J., and Xiao, C.
\newblock Constrained generation of semantically valid graphs via regularizing
  variational autoencoders.
\newblock In \emph{Advances in Neural Information Processing Systems}, pp.\
  7113--7124, 2018.

\bibitem[Mayr et~al.(2016)Mayr, Klambauer, Unterthiner, and
  Hochreiter]{mayr2016deeptox}
Mayr, A., Klambauer, G., Unterthiner, T., and Hochreiter, S.
\newblock Deeptox: toxicity prediction using deep learning.
\newblock \emph{Frontiers in Environmental Science}, 3:\penalty0 80, 2016.

\bibitem[Mueller et~al.(2017)Mueller, Gifford, and
  Jaakkola]{mueller2017sequence}
Mueller, J., Gifford, D., and Jaakkola, T.
\newblock Sequence to better sequence: continuous revision of combinatorial
  structures.
\newblock In \emph{International Conference on Machine Learning}, pp.\
  2536--2544, 2017.

\bibitem[Olivecrona et~al.(2017)Olivecrona, Blaschke, Engkvist, and
  Chen]{olivecrona2017molecular}
Olivecrona, M., Blaschke, T., Engkvist, O., and Chen, H.
\newblock Molecular de-novo design through deep reinforcement learning.
\newblock \emph{Journal of cheminformatics}, 9\penalty0 (1):\penalty0 48, 2017.

\bibitem[Popova et~al.(2018)Popova, Isayev, and Tropsha]{popova2018deep}
Popova, M., Isayev, O., and Tropsha, A.
\newblock Deep reinforcement learning for de novo drug design.
\newblock \emph{Science advances}, 4\penalty0 (7):\penalty0 eaap7885, 2018.

\bibitem[Putin et~al.(2018)Putin, Asadulaev, Ivanenkov, Aladinskiy,
  Sanchez-Lengeling, Aspuru-Guzik, and Zhavoronkov]{putin2018reinforced}
Putin, E., Asadulaev, A., Ivanenkov, Y., Aladinskiy, V., Sanchez-Lengeling, B.,
  Aspuru-Guzik, A., and Zhavoronkov, A.
\newblock Reinforced adversarial neural computer for de novo molecular design.
\newblock \emph{Journal of chemical information and modeling}, 58\penalty0
  (6):\penalty0 1194--1204, 2018.

\bibitem[Pyzer-Knapp et~al.(2015)Pyzer-Knapp, Suh, G{\'o}mez-Bombarelli,
  Aguilera-Iparraguirre, and Aspuru-Guzik]{pyzer2015high}
Pyzer-Knapp, E.~O., Suh, C., G{\'o}mez-Bombarelli, R., Aguilera-Iparraguirre,
  J., and Aspuru-Guzik, A.
\newblock What is high-throughput virtual screening? a perspective from organic
  materials discovery.
\newblock \emph{Annual Review of Materials Research}, 45:\penalty0 195--216,
  2015.

\bibitem[Reymond et~al.(2010)Reymond, Van~Deursen, Blum, and
  Ruddigkeit]{reymond2010chemical}
Reymond, J.-L., Van~Deursen, R., Blum, L.~C., and Ruddigkeit, L.
\newblock Chemical space as a source for new drugs.
\newblock \emph{MedChemComm}, 1\penalty0 (1):\penalty0 30--38, 2010.

\bibitem[Rezende et~al.(2014)Rezende, Mohamed, and
  Wierstra]{rezende2014stochastic}
Rezende, D.~J., Mohamed, S., and Wierstra, D.
\newblock Stochastic backpropagation and approximate inference in deep
  generative models.
\newblock In \emph{International Conference on Machine Learning}, pp.\
  1278--1286, 2014.

\bibitem[Rogers \& Hahn(2010)Rogers and Hahn]{rogers2010extended}
Rogers, D. and Hahn, M.
\newblock Extended-connectivity fingerprints.
\newblock \emph{Journal of chemical information and modeling}, 50\penalty0
  (5):\penalty0 742--754, 2010.

\bibitem[Rolfe(2016)]{rolfe2016discrete}
Rolfe, J.~T.
\newblock Discrete variational autoencoders.
\newblock \emph{arXiv preprint arXiv:1609.02200}, 2016.

\bibitem[Ryu et~al.(2018)Ryu, Lim, and Kim]{ryu2018deeply}
Ryu, S., Lim, J., and Kim, W.~Y.
\newblock Deeply learning molecular structure-property relationships using
  graph attention neural network.
\newblock \emph{arXiv preprint arXiv:1805.10988}, 2018.

\bibitem[Samanta et~al.(2018)Samanta, De, Jana, Chattaraj, Ganguly, and
  Gomez-Rodriguez]{samanta2018nevae}
Samanta, B., De, A., Jana, G., Chattaraj, P.~K., Ganguly, N., and
  Gomez-Rodriguez, M.
\newblock {NeVAE:} a deep generative model for molecular graphs.
\newblock \emph{arXiv preprint arXiv:1802.05283}, 2018.

\bibitem[Sanchez-Lengeling \& Aspuru-Guzik(2018)Sanchez-Lengeling and
  Aspuru-Guzik]{sanchez2018inverse}
Sanchez-Lengeling, B. and Aspuru-Guzik, A.
\newblock Inverse molecular design using machine learning: Generative models
  for matter engineering.
\newblock \emph{Science}, 361\penalty0 (6400):\penalty0 360--365, 2018.

\bibitem[Segler et~al.(2017)Segler, Kogej, Tyrchan, and
  Waller]{segler2017generating}
Segler, M.~H., Kogej, T., Tyrchan, C., and Waller, M.~P.
\newblock Generating focused molecule libraries for drug discovery with
  recurrent neural networks.
\newblock \emph{ACS central science}, 4\penalty0 (1):\penalty0 120--131, 2017.

\bibitem[Simonovsky \& Komodakis(2018)Simonovsky and
  Komodakis]{simonovsky2018graphvae}
Simonovsky, M. and Komodakis, N.
\newblock Graphvae: Towards generation of small graphs using variational
  autoencoders.
\newblock In \emph{International Conference on Artificial Neural Networks},
  pp.\  412--422. Springer, 2018.

\bibitem[Snelson \& Ghahramani(2006)Snelson and Ghahramani]{snelson2006sparse}
Snelson, E. and Ghahramani, Z.
\newblock Sparse gaussian processes using pseudo-inputs.
\newblock In \emph{Advances in neural information processing systems}, pp.\
  1257--1264, 2006.

\bibitem[Sterling \& Irwin(2015)Sterling and Irwin]{sterling2015zinc}
Sterling, T. and Irwin, J.~J.
\newblock Zinc 15--ligand discovery for everyone.
\newblock \emph{Journal of chemical information and modeling}, 55\penalty0
  (11):\penalty0 2324--2337, 2015.

\bibitem[Stumpfe \& Bajorath(2012)Stumpfe and Bajorath]{stumpfe2012exploring}
Stumpfe, D. and Bajorath, J.
\newblock Exploring activity cliffs in medicinal chemistry: miniperspective.
\newblock \emph{Journal of medicinal chemistry}, 55\penalty0 (7):\penalty0
  2932--2942, 2012.

\bibitem[Sutskever et~al.(2014)Sutskever, Vinyals, and
  Le]{sutskever2014sequence}
Sutskever, I., Vinyals, O., and Le, Q.~V.
\newblock Sequence to sequence learning with neural networks.
\newblock In \emph{Advances in neural information processing systems}, pp.\
  3104--3112, 2014.

\bibitem[Szegedy et~al.(2016)Szegedy, Vanhoucke, Ioffe, Shlens, and
  Wojna]{szegedy2016rethinking}
Szegedy, C., Vanhoucke, V., Ioffe, S., Shlens, J., and Wojna, Z.
\newblock Rethinking the inception architecture for computer vision.
\newblock In \emph{Proceedings of the IEEE conference on computer vision and
  pattern recognition}, pp.\  2818--2826, 2016.

\bibitem[Vinyals et~al.(2015)Vinyals, Bengio, and Kudlur]{vinyals2015order}
Vinyals, O., Bengio, S., and Kudlur, M.
\newblock Order matters: Sequence to sequence for sets.
\newblock \emph{arXiv preprint arXiv:1511.06391}, 2015.

\bibitem[Weininger(1988)]{weininger1988smiles}
Weininger, D.
\newblock {SMILES}, a chemical language and information system. 1.
  {I}ntroduction to methodology and encoding rules.
\newblock \emph{Journal of chemical information and computer sciences},
  28\penalty0 (1):\penalty0 31--36, 1988.

\bibitem[Winter et~al.(2019{\natexlab{a}})Winter, Montanari, No{\'e}, and
  Clevert]{winter2019learning}
Winter, R., Montanari, F., No{\'e}, F., and Clevert, D.-A.
\newblock Learning continuous and data-driven molecular descriptors by
  translating equivalent chemical representations.
\newblock \emph{Chemical Science}, 2019{\natexlab{a}}.

\bibitem[Winter et~al.(2019{\natexlab{b}})Winter, Montanari, Steffen, Briem,
  Noe, and Clevert]{winter2019efficient}
Winter, R., Montanari, F., Steffen, A., Briem, H., Noe, F., and Clevert, D.
\newblock Efficient multi-objective molecular optimization in a continuous
  latent space.
\newblock 2019{\natexlab{b}}.

\bibitem[Wu et~al.(2018)Wu, Ramsundar, Feinberg, Gomes, Geniesse, Pappu,
  Leswing, and Pande]{wu2018moleculenet}
Wu, Z., Ramsundar, B., Feinberg, E.~N., Gomes, J., Geniesse, C., Pappu, A.~S.,
  Leswing, K., and Pande, V.
\newblock Molecule{N}et: {A} benchmark for molecular machine learning.
\newblock \emph{Chemical science}, 9\penalty0 (2):\penalty0 513--530, 2018.

\bibitem[You et~al.(2018)You, Liu, Ying, Pande, and Leskovec]{you2018graph}
You, J., Liu, B., Ying, R., Pande, V., and Leskovec, J.
\newblock Graph convolutional policy network for goal-directed molecular graph
  generation.
\newblock \emph{arXiv preprint arXiv:1806.02473}, 2018.

\bibitem[Zaslavskiy et~al.(2019)Zaslavskiy, J{\'e}gou, Tramel, and
  Wainrib]{zaslavskiy2019toxicblend}
Zaslavskiy, M., J{\'e}gou, S., Tramel, E.~W., and Wainrib, G.
\newblock Toxicblend: Virtual screening of toxic compounds with ensemble
  predictors.
\newblock \emph{Computational Toxicology}, 10:\penalty0 81--88, 2019.

\bibitem[Zhou et~al.(2018)Zhou, Kearnes, Li, Zare, and
  Riley]{zhou2018optimization}
Zhou, Z., Kearnes, S., Li, L., Zare, R.~N., and Riley, P.
\newblock Optimization of molecules via deep reinforcement learning.
\newblock \emph{arXiv preprint arXiv:1810.08678}, 2018.

\end{thebibliography}
\bibliographystyle{icml2019}

\newpage

\begin{appendix}
  
  \section{Datasets} \label{sec:datasets}
  
  SMILES strings, as well as the true values of the log octanol-water partition coefficient (logP), molecular weight (MW), and the quantitative estimate of drug-likeness (QED), are computed using RDKit~\cite{landrum2006rdkit}.

  \begin{figure}[tbh]
    \centering
    \begin{tabular}{cp{1cm}c}
      \includegraphics[height=2cm,trim={1cm 3cm 1cm 3cm},clip]{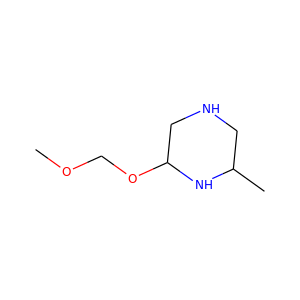} & &
      \includegraphics[height=2cm,trim={1cm 3cm 1cm 3cm},clip]{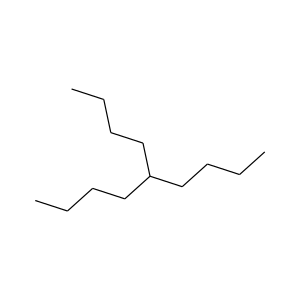} \\
      (a) COCOC1CNCC(C)N1 & & (b) CCCCC(CCCC)CCCC \\
      \hspace{0.4cm} CC1NC(CNC1)OCOC & &
    \end{tabular}
    \caption{Multiple SMILES strings of a single molecule may be more dissimilar than SMILES strings of radically dissimilar molecules. The top SMILES string for molecule~(a) is 30\% similar to the bottom SMILES string by string edit distance, but 60\% similar to the SMILES string for molecule~(b).}
    \label{fig:samemol_diffsmi}
  \end{figure}

  \subsection{ZINC}

\begin{figure}[tbh]
  \center
  \includegraphics[height=6cm,trim={10 5 40.0 40.0},clip]{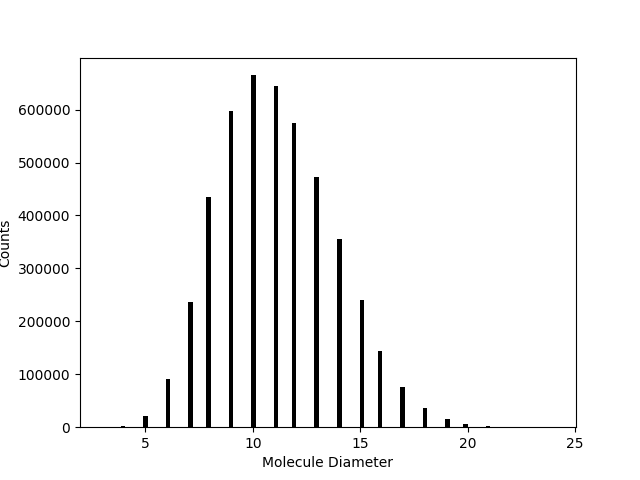}
  \caption{Histogram of molecular diameters in the ZINC250k dataset. The diameter is defined as the maximum eccentricity over all atoms in the molecular graph. The mean is 11.1; the maximum is 24. Typical implementations of graph convolution use only three to seven rounds of message passing~\cite{duvenaud2015convolutional, gilmer2017neural, jin2018junction, kearnes2016molecular, liu2018constrained, samanta2018nevae, you2018graph}, and so cannot propagate information across most molecules in this dataset.}
  \label{fig:graph_diam_hist}    
\end{figure}

For molecular property optimization and fully supervised property prediction, we train the All SMILES VAE on the ZINC250k dataset of 250,000 organic molecules with between 6 and 38 heavy atoms, and penalized logPs\footnote{The log octanol-water partition coefficient minus the synthetic accessibility score and the number of rings with more than six atoms.} from -13 to 5~\cite{gomez2018automatic}. This dataset is curated from the full ZINC12 dataset~\cite{irwin2012zinc}, and available from~\url{https://github.com/aspuru-guzik-group/chemical_vae}. The distribution of molecular diameters in ZINC250k is shown in Figure~\ref{fig:graph_diam_hist}.

For semi-supervised property prediction on logP, MW, and QED, we train on the ZINC310k dataset of 310,000 organic molecules with between 6 and 38 heavy atoms~\cite{kang2018conditional}. This dataset is curated from the full ZINC15 dataset~\cite{sterling2015zinc}, and available from~\mbox{\url{https://github.com/nyu-dl/conditional-molecular-design-ssvae}.}
  
\subsection{Tox21}

For the semi-supervised prediction of twelve forms of toxicity, we train on the Tox21 dataset~\cite{huang2016tox21challenge, mayr2016deeptox}, accessed through the DeepChem package~\cite{wu2018moleculenet}, with the provided random train/validation/test set split. This dataset contains binarized binding affinities against up to 12~proteins for 6264 training, 783 validation, and 784 test molecules. Tox21 contains molecules with up to 140~atoms, ranging from large peptides to lanthanide, actinide and other metals. Many of these metal atoms are not present in any of the standard molecular generative modeling datasets, and there are metal atoms in the validation and test set that never appear in the training set. To address these difficulties, we curated an unsupervised dataset of 1.5 million molecules from the PubChem database~\cite{kim2018pubchem}, which we will make available upon publication. 
To maintain commensurability with prior work, this additional unsupervised dataset is \emph{only} used on the Tox21 prediction task.

\section{Extended model architecture} \label{sec:extended_model_architecture}
\begin{figure}[tbh]
    \center
    \begin{tikzpicture}[->,>=stealth',shorten >=1pt,auto,node distance=1.5cm,semithick]
      \tikzstyle{every state}=[fill=white,draw=black,text=black,scale=0.7]
      \node[state]     (x1)                  {$\textbf{x}_1$};
      \node[state]     (x2) [below of=x1]    {$\textbf{x}_2$};
      \node[state]     (x3) [below of=x2]    {$\textbf{x}_3$};
      
      \node[draw=none] (x1_label) [left of=x1,xshift=1.6cm,yshift=0.5cm] {\tiny {\color{yellow} c}1{\color{red} c}({\color{green}Cl}){\color{purple} c}{\color{blue} n}{\color{pink} c}1};
      \node[draw=none] (x2_label) [left of=x2,xshift=1.6cm,yshift=0.5cm] {\tiny {\color{pink} c}1{\color{blue}n}{\color{purple} c}{\color{red} c}({\color{green} Cl}){\color{yellow} c}1};
      \node[draw=none] (x3_label) [left of=x3,xshift=1.6cm,yshift=0.5cm] {\tiny {\color{green} Cl}{\color{red} c}1({\color{purple} c}{\color{blue} n}{\color{pink} c}{\color{yellow} c}1)};
      
      \node[draw=black,rounded corners]     (RNN_1)        [right of=x1] {BiGRU};
      \node[draw=black,rounded corners]     (RNN_2)        [right of=x2] {BiGRU};
      \node[draw=black,rounded corners]     (RNN_3)        [right of=x3] {BiGRU};
      
      \node[draw=black,rounded corners]     (NN_1)        [right of=RNN_1] {Lin};
      \node[draw=black,rounded corners]     (NN_2)        [right of=RNN_2] {Lin};
      \node[draw=black,rounded corners]     (NN_3)        [right of=RNN_3] {Lin};
      
      \node[state]     (h11) [right of=NN_1]    {$\textbf{h}_1^1$};
      \node[state]     (h21) [right of=NN_2]    {$\textbf{h}_2^1$};
      \node[state]     (h31) [right of=NN_3]    {$\textbf{h}_3^1$};
      
      \node[draw=black,rounded corners]     (enc_1)        [right of=h21,xshift=-0.3cm] {$\text{Enc}^1$};
      
      \node[state]     (h22) [right of=enc_1]  {$\textbf{h}_2^2$};
      \node[state]     (h12) [above of=h22]    {$\textbf{h}_1^2$};
      \node[state]     (h32) [below of=h22]    {$\textbf{h}_3^2$};
      
      \node[draw=black,rounded corners]     (enc_2)        [right of=h22,xshift=-0.3cm] {$\text{Enc}^2$};
      
      \node[state]     (h23) [right of=enc_2]  {$\textbf{h}_2^3$};
      \node[state]     (h13) [above of=h23]    {$\textbf{h}_1^3$};
      \node[state]     (h33) [below of=h23]    {$\textbf{h}_3^3$};
      
      \node[draw=black,rounded corners]     (ap)        [right of=h23,xshift=-0.3cm] {AP};
      
      \node[state]     (z)   [right of=ap]     {$\textbf{z}$};
      
      \node[draw=black,rounded corners,scale=0.7]     (dec_2)        [right of=z]    {LSTM};
      \node[draw=black,rounded corners,scale=0.7]     (dec_1)        [above of=dec_2] {LSTM};
      \node[draw=black,rounded corners,scale=0.7]     (dec_3)        [below of=dec_2] {LSTM};
      
      \node[state]     (x1p) [right of=dec_1]    {$\textbf{x}_1'$};
      \node[state]     (x2p) [right of=dec_2]    {$\textbf{x}_2'$};
      \node[state]     (x3p) [right of=dec_3]    {$\textbf{x}_3'$};
      
      \node[draw=none] (x1p_label) [right of=x1p,xshift=-1.6cm,yshift=0.5cm] {\tiny {\color{green} Cl}{\color{red} c}1({\color{yellow} c}{\color{pink} c}{\color{blue} n}{\color{purple} c}1)};
      \node[draw=none] (x2p_label) [right of=x2p,xshift=-1.6cm,yshift=0.5cm] {\tiny {\color{purple} c}1{\color{blue} n}{\color{pink} c}{\color{yellow} c}{\color{red} c}1({\color{green} Cl})};
      \node[draw=none] (x3p_label) [right of=x3p,xshift=-1.6cm,yshift=0.5cm] {\tiny {\color{blue} n}1{\color{purple} c}{\color{red} c}({\color{green} Cl}){\color{yellow} c}{\color{pink} c}1};

      \path (x1)      edge[short color dash]  (RNN_1);
      \path (x2)      edge[short color dash]  (RNN_2);
      \path (x3)      edge[short color dash]  (RNN_3);
      \path (RNN_1)   edge[short color dash]  (NN_1);
      \path (RNN_2)   edge[short color dash]  (NN_2);
      \path (RNN_3)   edge[short color dash]  (NN_3);
      \path (NN_1)   edge[short color dash]   (h11);
      \path (NN_2)   edge[short color dash]   (h21);
      \path (NN_3)   edge[short color dash]   (h31);
      \path (h11)   edge[short color dash]    (enc_1);
      \path (h21)   edge[short color dash]    (enc_1);
      \path (h31)   edge[short color dash]    (enc_1);
      \path (enc_1)   edge[short color dash]   (h12);
      \path (enc_1)   edge[short color dash]   (h22);
      \path (enc_1)   edge[short color dash]   (h32);
      \path (h12)   edge[short color dash]    (enc_2);
      \path (h22)   edge[short color dash]    (enc_2);
      \path (h32)   edge[short color dash]    (enc_2);
      \path (enc_2)   edge[short color dash]   (h13);
      \path (enc_2)   edge[short color dash]   (h23);
      \path (enc_2)   edge[short color dash]   (h33);
      \path (h13)   edge[short color dash]    (ap);
      \path (h23)   edge[short color dash]    (ap);
      \path (h33)   edge[short color dash]    (ap);
      \path (ap)    edge[short color dash]    (z);
      \path (z)     edge[short color dash]   (dec_1);
      \path (z)     edge[short color dash]   (dec_2);
      \path (z)     edge[short color dash]   (dec_3);
      \path (dec_1)  edge[short color dash]   (x1p);
      \path (dec_2)  edge[short color dash]   (x2p);
      \path (dec_3)  edge[short color dash]   (x3p);

    \end{tikzpicture}
    \caption{Multiple SMILES strings representing a single, common molecule are preprocessed by a BiGRU and a linear transformation, followed by multiple encoder blocks as in Figures~\ref{fig:encoder_block} and~\ref{fig:atom_aggregation}. The approximating posterior depicted in Figure~\ref{fig:approx_post} then produces a sample from the latent state~$\mathbf{z}$, which is decoded into SMILES strings by LSTMs. Note that all SMILES strings, in both the input and the output, are distinct. The encoder blocks also receive a linear embedding of the original SMILES strings as input.}
    \label{fig:full_architecture}
  \end{figure}
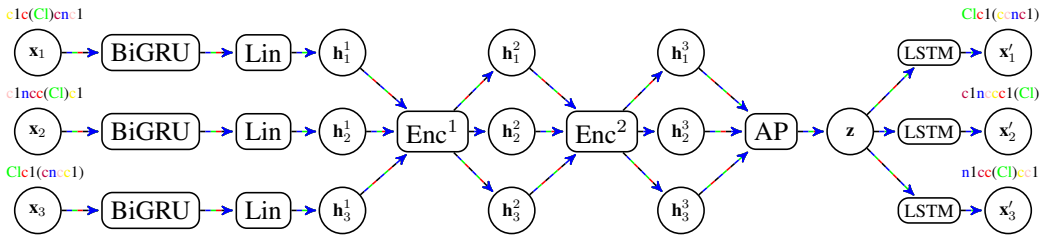

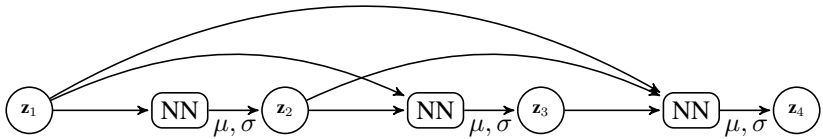
\begin{figure}[tbh]
    \center
    \begin{tikzpicture}[->,>=stealth',shorten >=1pt,auto,node distance=2.0cm,semithick]
      \tikzstyle{every state}=[fill=white,draw=black,text=black,scale=0.7]
      \node[state]                        (z1)                       {$\textbf{z}_1$};
      \node[draw=black,rounded corners]   (NN1)    [right of=z1]    {NN};
      \node[state]                        (z2)      [right of=NN1]  {$\textbf{z}_2$};
      \node[draw=black,rounded corners]   (NN2)    [right of=z2]    {NN};
      \node[state]                        (z3)      [right of=NN2]   {$\textbf{z}_3$};
      \node[draw=black,rounded corners]   (NN3)    [right of=z3]    {NN};
      \node[state]                        (z4)      [right of=NN3]   {$\textbf{z}_4$};
      
      \path (z1) edge (NN1);
      \path (NN1) edge node[below] {$\mu,\sigma$} (z2);
      \path (z1) edge[out=25,in=155] (NN2);
      \path (z2) edge (NN2);
      \path (NN2) edge node[below] {$\mu,\sigma$} (z3);
      \path (z1) edge[out=30,in=150] (NN3);
      \path (z2) edge[out=25,in=155] (NN3);
      \path (z3) edge (NN3);
      \path (NN3) edge node[below] {$\mu,\sigma$} (z4);
        
    \end{tikzpicture}
    \caption{The prior distribution over $\mathbf{z} = \left[\mathbf{z_1}, \mathbf{z_2}, \cdots \right]$ is a hierarchy of autoregressive Gaussians. The conditional prior distribution of hierarchical layer $i$ given layers $1$ through $i-1$, $p(\mathbf{z}_i | \mathbf{z}_1, \mathbf{z}_2, \cdots \mathbf{z}_{i-1})$, is a Gaussian with mean $\mu$ and log-variance $\log \sigma^2$ determined by a neural network with input $\left[ \mathbf{z}_1, \mathbf{z}_2, \cdots, \mathbf{z}_{i-1} \right]$.}
    \label{fig:autoregressive_prior}
  \end{figure}

The full All SMILES VAE architecture is summarized in Figure~\ref{fig:full_architecture}.
The evidence lower bound (ELBO) of the log-likelihood,
\begin{equation} \label{eq:elbo}
  \mathcal{L} = \mathbb{E}_{q(z|x)} \left[ \log p(x|z) \right] - \KL \left[ q(z|x) || p(z) \right],
\end{equation}
is the sum of the conditional log-likelihoods of $\mathbf{x}_i'$ in Figure~\ref{fig:full_architecture}, minus the Kullback-Leibler divergence between the approximating posterior, $q(z|x)$, computed by node $\text{AP}$ in Figure~\ref{fig:full_architecture}, and the prior depicted in Figure~\ref{fig:autoregressive_prior}.

  In all experiments, we use encoder stacks of depth three, with 512 hidden units in each GRU. The approximating posterior uses four layers of hierarchy, with 128 hidden units in the one-hidden-layer neural network that computes the attentional query vector. In practice, separate GRUs were used to produce the final hidden state for $\textbf{z}_1$ and the atom representations for $\textbf{z}_{>1}$. The single-layer LSTM decoder has 2048 hidden units. Training was performed using ADAM, with a decaying learning rate and 
  KL annealing. In all multiple SMILES strings architectures, we use 5 SMILES strings for encoding and decoding.

In contrast to many previous molecular VAEs, we do not scale down the term $\KL\left[ q(z|x) || p(z) \right]$ in the ELBO by the number of latent units~\cite{dai2018syntax,  kusner2017grammar}. However, our loss function does include separate reconstructions for multiple SMILES strings of a single molecule. 
For molecular optimization tasks, we usually scale up this KL term by the number of SMILES strings in the decoder, analogous to multiple single-SMILES VAEs in parallel; we leave the KL term unscaled for property prediction.

\subsection{Gated recurrent neural networks} \label{sec:GRU_review}

Convolutional neural networks on images~\cite{krizhevsky2012imagenet, lecun1990handwritten} 
leverage the inherent geometry of the visual domain to perform local message passing. At every spatial location of each layer, a convolutional network computes a message comprising the weighted sum of messages from the surrounding region in the preceding layer, followed by a point-wise nonlinearity. 
Each round of messages propagates information a distance equal to the convolutional kernel diameter multiplied by the current spatial resolution. 

Recurrent neural networks, such as long short-term memories~(LSTMs)~\cite{hochreiter1997long} and gated recurrent units~(GRUs)~\cite{cho2014learning}, model text, audio, and other one-dimensional sequences in an analogous manner. The kernel, comprising the weights on the previous hidden state and the current input, has a width of only two. However, the messages (i.e., the hidden states) are updated consecutively along the sequence, so information can propagate through the entire network in a single pass, substantially reducing the number of layers required. LSTMs and GRUs are ubiquitous in natural language processing tasks, and efficient GPU implementations have been developed~\cite{appleyard2016optimizing}.

Gated recurrent units (GRUs) are defined by~\cite{cho2014learning}:
\begin{align*}
[r, z] &= \sigma \left( x_t \left[ W_r, W_z \right] + h_{t-1} \left[ U_r, U_z \right] + [b_r, b_z] \right) \\
h_t &= (1 - z) \odot h_{t-1} + z \odot \tanh \left( x_t W + \left( r \odot h_{t-1} \right) U + b_h \right)
\end{align*}
where $r$, $z$, and $h$ are row-vectors, $[x, y]$ denotes the column-wise concatenation of $x$ and $y$, and the logistic function $\sigma(x) = \left(1 + e^{-x} \right)^{-1}$ and hyperbolic tangent are applied element-wise to vector argument $x$.
The hidden state $h_t$, comprising the message from node $t$, is a gated, weighted sum of the previous message $h_{t-1}$ and the current input $x_t$, both subject to an element-wise linear transformation and nonlinear (sigmoid) transformation. Specifically, the sum of the message from the input, $x_t W U^{-1}$, and the gated message from the previous node, $r \odot h_{t-1}$, is subject to a linear transformation $U$ and a pointwise nonlinearity. This is then gated and added to a gated residual connection from the previous node. 

Long short-term memories (LSTMs) are defined similarly~\cite{hochreiter1997long}:
\begin{align*}
[f_t, i_t, o_t] &= \sigma \left( x_t [W_f, W_i, W_o] + h_{t-1} [U_f, U_i, U_o] + [b_f, b_i, b_o] \right) \\
c_t &= f_t \odot c_{t-1} + i_t \odot \tanh \left( x_t W_c + h_{t-1} U_c + b_c \right) \\
h_t &= o_t \odot \tanh \left( c_t \right)
\end{align*}
where $f$ is the forget gate, $i$ is the input gate, and $o$ is the output gate.
LSTMs impose a second hyperbolic tangent and gating unit on the nonlinear recurrent message, but nevertheless still follow the form of applying width-two kernels and pointwise nonlinearities to the input and hidden state.

In contrast, message passing in graphs is defined by~\cite{duvenaud2015convolutional, kearnes2016molecular, kipf2016semi, li2015gated}: 
\begin{equation*}
  h_t^{(n)} = f \left( \left( \sum_{m \in \mathcal{N}(n)} h_{t-1}^{(m)} \right) W_t \right)
\end{equation*}
where $\mathcal{N}(n)$ is the set of neighbors of node $n$, for which there is an edge between $n$ and $m \in \mathcal{N}(n)$, and $f(x)$ is a pointwise nonlinearity such as a logistic function or rectified linear unit.
This message passing, also called graph convolutions, can be understood as a first-order approximation to spectral convolutions on graphs~\cite{hammond2011wavelets}.
\citet{kipf2016semi} additionally normalize each message by the square root of the degree of each node before and after the sum over neighboring nodes. \citet{kearnes2016molecular}
maintain separate messages for nodes and edges, with the neighborhood of a node comprising the connected edges and the neighborhood of an edge comprising the connected nodes. \citet{li2015gated} add gating analogous to a GRU.

An LSTM, taking a SMILES string as input, can realize a subset of the messages passed by graph convolutions.
For instance, input gates and forget gates can conspire to ignore open-parentheses, which indicate the beginning of a branch of the depth-first spanning tree traversal. If they similarly ignore the digits that close broken rings, the messages along each branch of the flattened spanning tree are not affected by the extraneous SMILES syntax. Input and forget gates can then reset the LSTM's memory at close-parentheses, which indicate the end of a branch of the depth-first spanning tree traversal, and the return to a previous node, ensuring that messages only propagate along connected paths in the molecular graph.

A set of LSTMs on multiple SMILES strings of a single molecule, with messages exchanged between the LSTMs, can generate all of the messages produced by a graph convolution. Atom-based pooling between LSTMs on multiple SMILES strings of the same molecule combines the messages produced in each flattened spanning tree, allowing every LSTM to access all messages produced by a graph convolution. While an LSTM decoder generating SMILES strings faces ambiguity regarding which set of SMILES strings representing a molecule to produce, this is analogous to the problem faced by graph-based decoders, as discussed in Appendix~\ref{sec:graph_redundancy}

\subsection{Bahdanau-style attention} \label{sec:bahdanau_attention}

The layers of the hierarchical approximating posterior after the first define a conditional Gaussian distribution, the mean and log-variance of which are parametrized by an attentional mechanism of the form proposed by~\citet{bahdanau2014neural}. The final encoder hidden vectors for each atom comprise the key vectors $k$, whereas the query vector $q$ is computed by a one-hidden-layer network of rectified linear units given the concatenation of the previous latent layers as input. The final output of the attentional mechanism, $c$, is computed via:
\begin{align*}
  e_i &= \tanh \left(q W_a + k_i U_a \right) v^{\top} \\
  \alpha_i &= \frac{\exp(e_i)}{\sum_j \exp(e_j)} \\
  c &= \sum_i \alpha_i k_i 
\end{align*}
The output of the attentional mechanism is subject to batch renormalization and a linear transformation to compute the conditional mean and log-variance of the layer.

\subsection{Latent space optimization}

To further ensure that the optimization is constrained to well-trained regions of latent space, we add~$\lambda \log (p(z))$ to the objective function, where~$\lambda$ is a hyperparameter.
Finally, to moderate the strictly monotonic nature of linear regressors, we apply an element-wise hard tanh to all latent variables before the regressor, with a linear region that encompasses all values observed in the training set.

To compare with previous work as fairly as possible, we optimize 1000 random samples from the prior to convergence, collecting the last point from each trajectory with a valid SMILES decoding. From these 1000 points, we evaluate the true molecular property on the 100 points for which the predicted property value is largest. Of these 100 values, we report the three largest. However, optimization within our latent space is computationally inexpensive, and requires no additional property measurement data. We could somewhat improve molecular optimization at minimal expense by constructing additional optimization trajectories in latent space, and evaluating the true molecular properties on the best points from this larger set.

Molecular optimization is quite robust to hyperparameters. We considered ADAM learning rates in $\left\{ 0.1, 0.01, 0.001, 0.0001 \right\}$ and $\lambda \in \left\{ 0.1, 0.01, 0.001, 0.0001 \right\}$. 

\subsection{Summary of novel contributions}

Starting with the work of \citet{gomez2018automatic}, previous molecular variational autoencoders have used one consistent SMILES string as both the input to the RNN encoder and the target of the RNN decoder. Any single SMILES string explicitly represents only a subset of the pathways in the molecular graph. Correspondingly, the recurrent neural networks in these encoders implicitly propagated information through only a fraction of the possible pathways. \citet{kipf2016variational}, 
\citet{liu2018constrained}, and \citet{simonovsky2018graphvae}, amongst others, trained molecular VAEs with graph convolutional encoders, which pass information through all graph pathways in parallel, but at considerable computational expense. None of these works used enough layers of graph convolutions to transfer information across the diameter of the average molecule in standard drug design datasets. The All SMILES VAE introduces the use of multiple SMILES strings of a single, common molecule as input to a RNN encoder, with pooling of homologous messages amongst the hidden representations associated with different SMILES strings. This allows information to flow through all pathways of the molecular graph, but can efficiently propagate information across the entire width of the molecule in a single layer.

\citet{bjerrum2018improving} and \citet{winter2019learning} trained sequence-to-sequence transcoders to map between different SMILES strings of the same molecule. These transcoders do not define an explicit generative model over molecules, and their latent spaces have no prior distributions. The All SMILES VAE extends this approach to variational autoencoders, and thereby learns a SMILES-derived generative model of molecules, rather than SMILES strings. The powerful, learned, hierarchical prior of the All SMILES VAE regularizes molecular optimization and property prediction. To ensure that molecular property optimization searches within the practical support of the prior, containing almost all of its probability mass, we introduce a hierarchical radius constraint on optimization with respect to the latent space.

\section{Extended results} \label{sec:extended_results}

We compare the performance of the All SMILES VAE to a variety of state-of-the-art algorithms that have been evaluated on standard molecular property prediction and optimization tasks. In particular, we compare to previously published results on the character/chemical VAE (CVAE)~\cite{gomez2018automatic} (with results reported in~\cite{kusner2017grammar}), grammar VAE (GVAE)~\cite{kusner2017grammar}, syntax-directed VAE (SD-VAE)~\cite{dai2018syntax}, junction tree VAE (JT-VAE)~\cite{jin2018junction}, NeVAE~\cite{samanta2018nevae}, semisupervised VAE (SSVAE)~\cite{kang2018conditional}, graph convolutional policy network (GCPN)~\cite{you2018graph}, molecule deep Q-network (MolDQN)~\cite{zhou2018optimization}, and the DeepChem~\cite{wu2018moleculenet} implementation of extended connectivity fingerprints (ECFP)~\cite{rogers2010extended} and graph convolutions (GraphConv)~\cite{duvenaud2015convolutional, kearnes2016molecular, wu2018moleculenet}. Extended connectivity fingerprints are a fixed-length hash of local fragments of the molecule, used as input to conventional machine learning techniques such as random forests, support vector machines, and non-convolutional neural networks~\cite{wu2018moleculenet}. For toxicity prediction, we also compare to 
PotentialNet~\cite{feinberg2018potentialnet}, ToxicBlend~\cite{zaslavskiy2019toxicblend}, and the results of \citet{li2017learning}. 

\subsection{Reconstruction accuracy and validity}

Previous molecular variational autoencoders have been evaluated using the percentage of molecules that are correctly reconstructed when sampling from both the approximating posterior $q(z|x)$ and the conditional likelihood $p(x|z)$ (reconstruction accuracy), and the percentage of samples from the prior $p(z)$ and conditional likelihood $p(x|z)$ that are valid SMILES strings (validity).
While these measure have intuitive appeal, they reflect neither the explicit training objective (the ELBO), nor the requirements of molecular optimization. In particular, when optimizing molecules via the latent space, a deterministic decoder ensures that each point in latent space is associated with a single set of well-defined molecular properties. 

The All SMILES VAE is trained on a more difficult task than previous molecular VAEs, since the reconstruction targets are different SMILES encodings than those input to the approximating posterior. This ensures that the latent representation only captures the molecule, rather than its particular SMILES encoding, but it requires the decoder LSTM to produce a complex, highly multimodal distribution over SMILES strings. As a result, samples from the decoder distribution are less likely to correspond to the input to the encoder, either due to syntactic or semantic errors.

To compensate for this unusually difficult decoding task, we evaluate the All SMILES VAE using a beam search over the decoder distribution.\footnote{The full decoder distribution is still used for training.} That is, we decode to the single SMILES string estimated to be most probable under the conditional likelihood $p(x|z)$. This has the added benefit of defining an unambiguous decoding for every point in the latent space, simplifying the interpretation of optimization in the latent space (Section~\ref{sec:molecular_optimization}). However, it renders the reconstruction and validity reported in Section~\ref{sec:results} incommensurable with much prior work, which use stochastic encoders and decoders.

\subsection{Property prediction}

Rather than jointly modeling the space of molecules and their properties, some earlier molecular variational autoencoders
first trained an unsupervised VAE on molecules, extracted their latent representations, and then trained a sparse Gaussian process over molecular properties as a function of these fixed latent representations~\cite{dai2018syntax, jin2018junction, kusner2017grammar, samanta2018nevae}.
Sparse Gaussian processes are parametric regressors, with the location and value of the inducing points trained based upon the entire supervised dataset~\cite{snelson2006sparse}. They have significantly more parameters, and are corresponding more powerful, than linear regressors.

Molecular properties are only a smooth function of the VAE latent space when the property regressor is trained jointly with the generative model~\cite{gomez2018automatic}. Results using a sparse Gaussian process on the latent space of an unsupervised VAE are very poor compared to less powerful regressors trained jointly with the VAE. Our property prediction is two orders of magnitude more accurate than sparse Gaussian process regression on an unsupervised VAE latent representation, as shown in Table~\ref{tbl:supervised_logp_with_pretrained_generative_model}. 

\begin{table}[tbh]
  \caption{
    Root-mean-square error of the log octanol-water partition coefficient (logP) on the ZINC250k dataset. Results other than the All SMILES VAE are those reported in the cited papers.}
  \label{tbl:supervised_logp_with_pretrained_generative_model}
  \begin{center}
      \begin{sc}
        \begin{tabular}{lll}
          \toprule
          Model & RMSE    \\ 
          \midrule
          Character VAE (CVAE)~\cite{gomez2018automatic, kusner2017grammar} & 1.504 \\ 
          Grammar VAE (GVAE)~\cite{kusner2017grammar} & 1.404 \\ 
          Syntax-directed VAE (SD-VAE)~\cite{dai2018syntax} & 1.366 \\ 
          Junction tree VAE (JT-VAE)~\cite{jin2018junction} & 1.290 \\ 
          NeVAE~\cite{samanta2018nevae} & 1.23 \\ 
          {\bf All SMILES} & {\bf 0.011 $\pm$ 0.001} \\ 
          \bottomrule
        \end{tabular}
      \end{sc}
  \end{center}
\end{table}

We report numerical results on semi-supervised property prediction, as well as comparisons from \citet{kang2018conditional}, in Table~\ref{tbl:semi_supervised_logp}. Our mean absolute error is at least three times smaller than comparison algorithms on the log octanol-water partition coefficient (logP) and molecular weight (MW). 

\begin{table}[tbh]
  \caption{Mean absolute error (MAE) of semi-supervised property prediction on the log octanol-water partition coefficient (logP), molecular weight (MW), and the quantitative estimate of drug-likeness (QED) on ZINC310k dataset. Results other than the All SMILES VAE are those reported by~\cite{kang2018conditional}.}
  \label{tbl:semi_supervised_logp}
  \begin{center}
      \begin{sc}
        \begin{tabular}{lllll}
          \toprule
          Model & \% labeled & MAE logP & MAE MW & MAE QED    \\
          \midrule
          ECFP      &  50\% & 0.180 $\pm$ 0.003  & 9.012 $\pm$ 0.184 & 0.023 $\pm$ 0.000 \\
          GraphConv &  50\% & 0.086 $\pm$ 0.012  & 4.506 $\pm$ 0.279 & 0.018 $\pm$ 0.001 \\ 
          SSVAE     &  50\% & 0.047  $\pm$ 0.003 & 1.05  $\pm$ 0.164 & 0.01  $\pm$ 0.001  \\
          All SMILES & 50\% & 0.007 $\pm$ 0.002  & 0.21 $\pm$ 0.07   & 0.0064 $\pm 0.0002$ \\
          \midrule
          ECFP      &  20\% & 0.249 $\pm$ 0.004  & 12.047 $\pm$ 0.168 & 0.033 $\pm$ 0.001 \\
          GraphConv &  20\% & 0.112 $\pm$ 0.015  & 4.597 $\pm$ 0.419 & 0.021 $\pm$ 0.002 \\ 
          SSVAE     &  20\% & 0.071 $\pm$ 0.007  & 1.008 $\pm$ 0.370 & 0.016 $\pm$ 0.001 \\
          All SMILES & 20\% &  0.009 $\pm$ 0.002             &0.33 $\pm$0.06             & 0.0079 $\pm 0.0003$ \\
          \midrule
          ECFP      &  10\% & 0.335 $\pm$ 0.005  & 15.057 $\pm$ 0.358 & 0.045 $\pm$ 0.001 \\
          GraphConv &  10\% & 0.148 $\pm$ 0.016  & 5.255 $\pm$ 0.767 & 0.028 $\pm$ 0.003 \\ 
          SSVAE &  10\% & 0.090  $\pm$ 0.004 & 1.444 $\pm$ 0.618 & 0.021 $\pm$ 0.001  \\
          All SMILES & 10\% & 0.014 $\pm$ 0.002              & 0.30 $\pm$ 0.06           & 0.0126 $\pm$ 0.0006 \\
          \midrule
          ECFP      &   5\% & 0.380 $\pm$ 0.009  & 17.713 $\pm$ 0.396 & 0.053 $\pm$ 0.001 \\
          GraphConv &   5\% & 0.187 $\pm$ 0.015  & 6.723 $\pm$ 2.116 & 0.034 $\pm$ 0.004 \\ 
          SSVAE &   5\% & 0.120  $\pm$ 0.006 & 1.639 $\pm$ 0.577 & 0.028 $\pm$ 0.001  \\
          All SMILES &  5\% & 0.036 $\pm$ 0.004               & 0.4 $\pm$ 0.1             & 0.0217 $\pm$ 0.0003  \\
          \bottomrule
        \end{tabular}
      \end{sc}
  \end{center}
\end{table}

\begin{figure}[tbh]
  \center
  \subfloat[True logP]{ \label{fig:true_heat_map}
    \includegraphics[height=4.7cm,trim={1.3cm 0.6cm 1.6cm 1.4cm},clip]{logp_heatmap.png}
  }
  \subfloat[Predicted logP]{ \label{fig:pred_heat_map}
    \includegraphics[height=4.7cm,trim={1.3cm 0.6cm 4cm 1.4cm},clip]{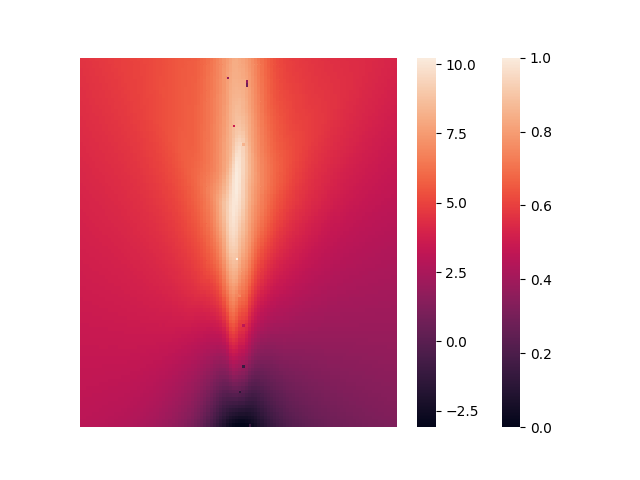}
  }
  \caption{Dense decodings of true logP (a) and predicted logP (b) along a local 2D sheet in latent space, with the y axis aligned with the trained logP regressor.}
  \label{fig:smoothness_pred_v_true}
\end{figure}

As a visual demonstration of the accuracy of property prediction, in Figure~\ref{fig:smoothness_pred_v_true} we show the predicted logP of a 2D slice of latent space subject to the hierarchical radius constraint, alongside the true logP of the molecules decoded from this slice (identical to Figure~\ref{fig:heat_map}).

Pathways on which activity (active or inactive) is assessed for the Tox21 dataset include seven nuclear receptor signaling pathways: androgen receptor, full (NR-AR)
androgen receptor, LBD (NR-AR-LBD);
aryl hydrocarbon receptor (NR-AHR);
aromatase (NR-AROMATASE);
estrogen receptor alpha, LBD (NR-ER-LBD);
estrogen receptor alpha, full (NR-ER);
and peroxisome proliferator-activated receptor gamma (NR-PPAR-GAMMA).
The Tox21 dataset also includes activity assessments for five stress response pathways:
nuclear factor (erythroid-derived 2)-like 2/antioxidant responsive element (SR-ARE);
ATAD5 (SR-ATAD5);
heat shock factor response element (SR-HSE);
mitochondrial membrane potential (SR-MMP);
and p53 (SR-p53).
We report the area under the receiver operating characteristic curve (AUC-ROC) on each assay independently in Table~\ref{tbl:per_assay_auc}. The average of these AUC-ROCs is reported in Table~\ref{tbl:fully_supervised}. We do not include the result of \citet{kearnes2016molecular} in Table~\ref{tbl:fully_supervised}, since it is not evaluated on the same train/validation/test split of the Tox21 dataset, and so is not commensurable.

\begin{table}[tbh]
  \caption{Area under the receiver operating characteristic curve (AUC-ROC) per assay on the Tox21 dataset.}
  \label{tbl:per_assay_auc}
  \begin{tabular}{llllll}
	\toprule
	{\small NR-AR} & {\small NR-AR-LBD} & {\small NR-AHR} & {\small NR-AROMATASE} & {\small NR-ER} & {\small NR-ER-LBD} \\
	\midrule
	0.868 & 0.907 & 0.889 & 0.907 & 0.714 & 0.830 \\
        \toprule
        {\small NR-PPAR-GAMMA} & {\small SR-ARE} & {\small SR-ATAD5} & {\small SR-HSE} & {\small SR-MMP} & {\small SR-p53} \\
        \midrule
        0.911 & 0.863 & 0.870 & 0.901 &0.914 & 0.888 \\
	\bottomrule
  \end{tabular}
\end{table}

\subsection{Molecular optimization} \label{sec:extended_molecular_optimization}

\begin{figure}[tbh]
  \centering
\includegraphics[height=5cm,trim={0cm 0cm 1cm 1.2cm},clip]{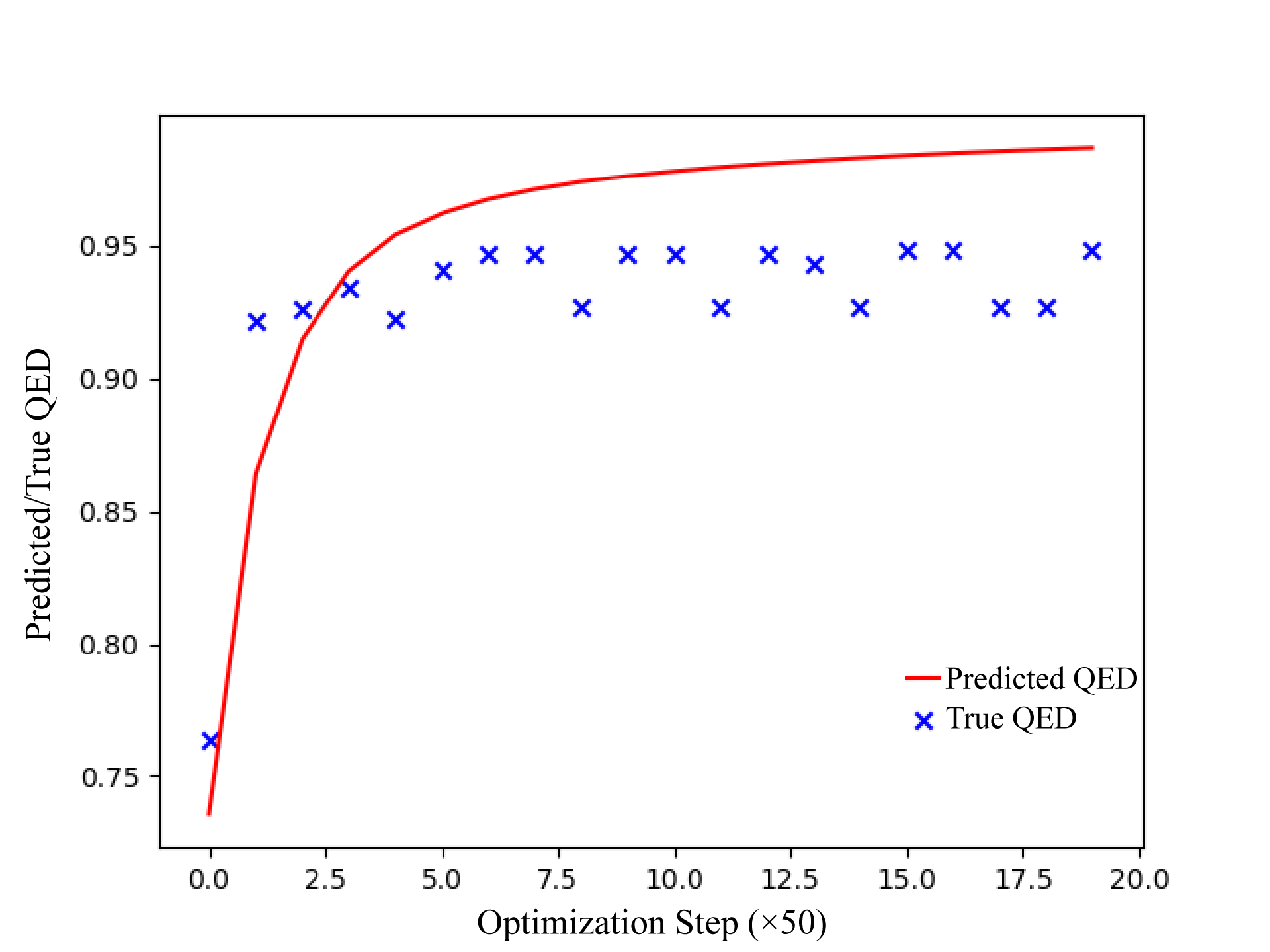}
\caption{Predicted (red line) and true (blue x's) quantitative estimate of drug-likeness (QED) over the optimization trajectory resulting in the molecule with the maximum observed true QED  (0.948).}
\label{fig:qed_optimization_trajectory}
\end{figure}

\begin{figure}[tbh]
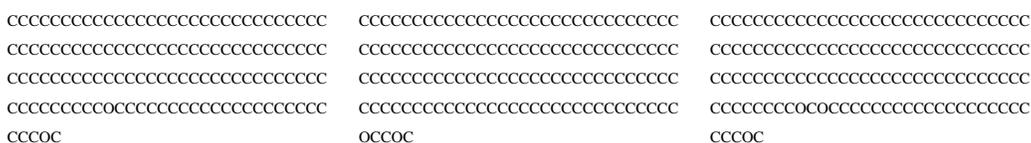

  \centering
  \begin{tabular}{p{4.25cm}p{4.25cm}p{4.25cm}}
    {\tiny \seqsplit{CCCCCCCCCCCCCCCCCCCCCCCCCCCCCCCCCCCCCCCCCCCCCCCCCCCCCCCCCCCCCCCCCCCCCCCCCCCCCCCCCCCCCCCCCCCCCCCCCCCOCCCCCCCCCCCCCCCCCCCCCCCOC}} & 
    {\tiny \seqsplit{CCCCCCCCCCCCCCCCCCCCCCCCCCCCCCCCCCCCCCCCCCCCCCCCCCCCCCCCCCCCCCCCCCCCCCCCCCCCCCCCCCCCCCCCCCCCCCCCCCCCCCCCCCCCCCCCCCCCCCCCOCCOC}} &
    {\tiny \seqsplit{CCCCCCCCCCCCCCCCCCCCCCCCCCCCCCCCCCCCCCCCCCCCCCCCCCCCCCCCCCCCCCCCCCCCCCCCCCCCCCCCCCCCCCCCCCCCCCCCCCOCOCCCCCCCCCCCCCCCCCCCCCCOC}} \\

  \end{tabular}
  \caption{Molecules with the top three true penalized LogP values produced by gradient-based optimization subject to the hierarchical radius constraint in the All SMILES VAE, but with the KL term unscaled by the number of SMILES strings in the decoder. Molecules are shown as SMILES strings, wrapped across multiple lines, as they are too large to be properly rendered into an image.}
  \label{fig:top_penlogp_lowkl}
\end{figure}

We present an optimization trajectory for the quantitative estimate of drug-likeness (QED) in Figure~\ref{fig:qed_optimization_trajectory}. For the molecules depicted in Figure~\ref{fig:top_molecules}, we scaled $\KL(q(z|x) || p(z))$) in the ELBO (Equation~\ref{eq:elbo}) of the All SMILES VAE by the number of SMILES strings in the decoder. This renders the loss function analogous to that of many parallel single-SMILES VAEs, but with message passing between encoders leading to a shared latent representation. If we leave the KL term unscaled, latent space embeddings are subject to less regularization forcing them to match the prior distribution. Optimization of molecular properties with respect to the latent space therefore searches over a wider space of molecules, which are less similar to the training set. In Figure~\ref{fig:top_penlogp_lowkl}, we show that such an optimization for penalized log P finds very long aliphatic chains, with penalized log P values as large as $42.46$.

\subsection{Ablation of model components} \label{sec:ablation}

The All SMILES VAE passes messages along and between the flattened spanning trees of molecular graphs encoded by multiple SMILES strings of a shared molecule. In the base implementation, this is realized via a stack of GRUs, with pooling across SMILES strings amongst the hidden representation of each atom in each layer, and further max pooling of the final hidden states. 
In Table~\ref{tbl:ablation}, we progressively ablate model components to demonstrate that this computational architecture builds a powerful fixed-length representation of molecules, rather than their particular SMILES string instantiations. We evaluate the effect of these model ablations on the mean absolute error of predictions of the log octanol-water partition coefficient (logP) and the quantitative estimate of drug-likeness (QED), as well as the percentage of samples from the prior that decode to valid SMILES strings (Val) and the percentage of molecules in a held-out test set that are reconstructed accurately by the encoder and decoder (Rec acc). In all cases, we use the layer-wise maximum a posterior configuration of the encoder (the mean of each conditional Gaussian distribution), and a beam-search decoder.

{\sc No atom-based pooling} removes the pooling between each instance of an atom across SMILES strings, depicted in Figure~\ref{fig:atom_aggregation}. As a result, the multiple SMILES inputs are processed independently until the final max pooling over GRU hidden states, which serves as input to the first layer of the latent variable hierarchy. A random SMILES string is chosen to serve as input to the attentional mechanism for the remaining layers of the latent variable hierarchy. The effect of this ablation on toxicity prediction for the Tox21 dataset is reported in Table~\ref{tbl:ablation_tox21}. We extend this process in {\sc One SMILES enc} by only feeding a single SMILES string to the encoder, although the decoder still reconstructs multiple disjoint SMILES strings. {\sc One SMILES enc/dec ($\neq$)} further reduces the size of the decoder set to one, but the encoded and decoded SMILES strings are distinct representations of the molecule. Finally, {\sc One SMILES enc/dec ($=$)} encodes and decodes a single, shared SMILES string.

Except for {\sc One SMILES enc/dec ($=$)}, all of these ablations primarily disrupt the flow of messages between the flattened spanning trees, and induce a similar, significant decay in performance. {\sc One SMILES enc/dec ($=$)} further permits the latent representation to encode the details of the particular SMILES string, rather than forcing the representation of only the underlying molecule, and causes a further reduction in performance.

We also observe a meaningful contribution from the hierarchical approximating posterior, with its attentional pooling mechanism.
In {\sc No posterior hierarchy}, we move all latent variables to the first layer of the hierarchy, removing the succeeding layers. The remaining prior is a standard Gaussian, and there is no attentional pooling over the atom representations.

Table~\ref{tbl:ablation_opt} shows that the hierarchical radius constraint significantly improves molecular optimization.
In contrast to Table~\ref{table:optimization_results}, optimization is performed on penalized logP alone, without a log prior regularizer. This produces better results without the radius constraint, and so constitutes a more conservative ablation experiment.

\begin{table}[tbh]
  \caption{Effect of model ablation on fully supervised property prediction and generative modeling using the ZINC250k dataset.}
  \label{tbl:ablation}
  \centering
  {\small
  \begin{sc}
    \begin{tabular}{lllll}
      \toprule
      Ablation & MAE logP & MAE QED & Val & Rec acc \\
      \midrule
      Full model & 0.005 $\pm$ 0.0006  & 0.0052 $\pm$ 0.0001 & 98.5 $\pm$ 0.1 & 87.4 $\pm$ 1.0 \vspace{0.15cm} \\ 
      No atom-based pooling & 0.008 $\pm$ 0.004  & 0.0076 $\pm$ 0.0005 & 97.6 $\pm$ 0.2 & 84.0 $\pm$ 0.4 \\
      One SMILES enc & 0.008 $\pm$ 0.0005  & 0.0073 $\pm$ 0.0002 & 98.4 $\pm$ 0.1 & 82.3 $\pm$ 0.4\\
      One SMILES enc/dec ($\neq$)   & 0.009 $\pm$ 0.001    & 0.0091 $\pm$ 0.0003 \\
      One SMILES enc/dec ($=$) & 0.025 $\pm$ 0.003 & 0.0115 $\pm$ 0.0004 \vspace{0.15cm} \\
      No posterior hierarchy & 0.010 $\pm$ 0.003 &  0.0051 $\pm$ 0.0001& \\
      \bottomrule
    \end{tabular}
  \end{sc}
}
\end{table}

\begin{table}[tbh]
  \caption{Effect of model ablation on Tox21 toxicity prediction.}
  \label{tbl:ablation_tox21}
  \centering
  \begin{sc}
    \begin{tabular}{ll}
      \toprule
      Ablation & AUC-ROC    \\
      \midrule
      Full model &  0.871 \\
      No atom-based pooling & 0.864 \\ 
      \bottomrule
    \end{tabular}
  \end{sc}
\end{table}

\begin{table}[tbh]
  \caption{Effect of the hierarchical radius constraint on penalized logP optimization. Predicted penalized logP was evaluated on 1000 optimization trajectories. From these, the true logP was evaluated on the 100 best trajectories, and the top three true penalized logPs are reported.
    Each optimization was repeated 5 times. }
  \label{tbl:ablation_opt}
  \centering
  \begin{sc}
    \begin{tabular}{llll}
      \toprule
      Ablation & 1st best logP & 2nd best logP & 3rd best logP    \\
      \midrule
      With radius constraint & 17.0 $\pm$ 3.0 & 16.0 $\pm$ 2.0 & 14.8 $\pm$ 0.3  \\
      Without radius constraint & 8.5044 $\pm$ 0.0  &6.9526 $\pm$ 0 & 5.36 $\pm$ 0.05\\
      \bottomrule
    \end{tabular}
  \end{sc}
\end{table}

\section{SMILES grammar can be enforced with a pushdown automaton} \label{sec:SMILES_grammar}

The subset of the SMILES grammar~\cite{weininger1988smiles} captured by \citet{dai2018syntax} and \citet{kusner2017grammar} is equivalent to the context-free grammar shown in Figure~\ref{fig:SMILES_grammar}.
This subset does not include the ability to represent multiple disconnected molecules in a single SMILES string, multiple fragments that are only connected by ringbonds, or wildcard atoms. \texttt{element\_symbols} includes symbols for every element in the periodic table, including the \texttt{aliphatic\_organic} symbols.

\begin{figure}[tbh]
  \begin{align*}
    \text{chain} &\rightarrow \text{branched\_atom rest\_of\_chain} \\
    \text{rest\_of\_chain} &\rightarrow \text{$\epsilon$ | bond? chain} \\
    \text{bond} &\rightarrow \text{`-' | `=' | `\#' | `\$' | `:' | `/' | `\textbackslash' } \\
    \text{branched\_atom} &\rightarrow \text{atom ringbond* branch*} \\
    \text{ringbond} &\rightarrow \text{bond digit? digit} \\
    \text{branch} &\rightarrow \text{`(' bond? chain `)'} \\
    \text{atom} &\rightarrow \text{aliphatic\_organic | aromatic\_organic | bracket\_atom} \\
    \text{aliphatic\_organic} &\rightarrow \text{`B' | `C' | `N' | `O' | `S' | `P' | `F' | `Cl' | `Br' | `I'} \\
    \text{aromatic\_organic} &\rightarrow \text{`b' | `c' | `n' | `o' | `s' | `p'} \\
    \text{bracket\_atom} &\rightarrow \text{`[' isotope? symbol chiral? hcount? charge? class? `]'} \\
    \text{isotope} &\rightarrow \text{digit? digit? digit} \\
    \text{symbol} &\rightarrow \text{element\_symbols | aromatic\_symbols} \\
    \text{aromatic\_symbols} &\rightarrow \text{`c' | `n' | `o' | `p' | `s' | `se' | `as'} \\
    \text{chiral} &\rightarrow \text{`@' | `@@' | `@TH1' | `@TH2' | `@AL1' | `@AL2' |} \\
    &\qquad \text{`@SP1' | `@SP2' | `@SP3' | `@TB1' | `@TB2' $\cdots$ `@TB30' | } \\
    &\qquad \text{`@OH1' | `@OH2' $\cdots$ `@OH30'} \\
    \text{hcount} &\rightarrow \text{`H' digit?} \\
    \text{charge} &\rightarrow \text{`-' digit? | `+' digit? }\\
    \text{class} &\rightarrow \text{`:' digit? digit? digit? } \\
    \text{digit} &\rightarrow \text{`0' | `1' | `2' | `3' | `4' | `5' | `6' | `7' | `8' | `9'}
  \end{align*}
  \caption{Context-free grammar of SMILES strings}
  \label{fig:SMILES_grammar}
\end{figure}

Productions generally begin with a unique, defining symbol or set of symbols. Exceptions include \texttt{bond} and \texttt{charge} (both can begin with \texttt{-}), and \texttt{aromatic\_organic} and \texttt{aromatic\_symbols} (both include \texttt{c}, \texttt{n}, \texttt{o}, \texttt{s}, and \texttt{p}), but these pairs of productions never occur in the same context, and so cannot be confused. 
The particular production for \texttt{chiral} can only be resolved by parsing characters up to the next production, but the end of \texttt{chiral} and the identity of the subsequent production can be inferred from its first symbol of the production after \texttt{chiral}. Alternatively, the strings of \texttt{chiral} can be encoded as monolithic tokens.

Whenever there is a choice between productions, the true production is uniquely identified by the next symbols.
The only aspect of the SMILES grammar that requires more than a few bits of memory is the matching of parentheses, which can be performed in a straightforward manner with a pushdown automaton.
As a result, parse trees~\cite{dai2018syntax, kusner2017grammar} need not be explicitly constructed by the decoder to enforce the syntactic restrictions of SMILES strings. Rather, the SMILES grammar can be enforced with a pushdown automaton running in parallel with the decoder RNN. The state of the pushdown automaton tracks progress within the representation of each atom, and the sequence of atoms and bonds. The set output symbols available to the decoder RNN is restricted to those consistent with the current state of the pushdown automaton. 
\texttt{(} and \texttt{[} are pushed onto the stack when are emitted, and must be popped from the top of the stack in order to emit \texttt{)} or \texttt{]} respectively.

For example, in addition to simple aliphatic organic (\texttt{B}, \texttt{C}, \texttt{N}, \texttt{O}, \texttt{S}, \texttt{P}, \texttt{F}, \texttt{Cl}, \texttt{Br}, or \texttt{I}) or aromatic organic (\texttt{b}, \texttt{c}, \texttt{n}, \texttt{o}, \texttt{s}, or \texttt{p}) symbols, an \texttt{atom} may be represented by a pair of brackets (requiring parentheses matching) containing a sequence of isotope number, atom symbol, chiral symbol, hydrogen count, charge, and class. With the exception of the atom symbol, each element of the sequence is optional, but is easily parsed by a finite state machine. 
\texttt{isotope}, \texttt{symbol}, \texttt{chiral}, \texttt{hcount}, \texttt{charge}, and \texttt{class} can all be distinguished based upon their first character, so the position in the progression can be inferred trivially.\footnote{\texttt{symbol} and \texttt{hcount} can both start with `H', but symbol is mandatory, so there is no ambiguity.}

When parsing \texttt{branched\_atom}, all productions after the initial \texttt{atom} are \texttt{ringbonds} until the first \texttt{(}, which indicates the beginning of a \texttt{branch}. After observing a \texttt{)}, and popping the complementary \texttt{(} off of the stack, the SMILES string is necessarily in the third component of a \texttt{branched\_atom}, since only a \texttt{branched\_atom} can emit a \texttt{branch}, and only \texttt{branch} produces the symbol \texttt{)}. The next symbol must be a \texttt{(}, indicating the beginning of another \texttt{branch}, or one of the first symbols of \texttt{rest\_of\_chain}, since this must follow the \texttt{branched\_atom} in the \texttt{chain} production.

\subsection{Ringbond and valence shell semantic constraints} \label{sec:valence_shell}

Similarly, the semantic restrictions of ringbond matching and valence shell constraints can be enforced during feedforward production of a SMILES string using a pushdown stack and a small (100-element) random access memory.
Our approach depends upon the presence of matching bond labels at both sides of a ringbond, which is allowed but not required in standard SMILES syntax. We assume the trivial extention of the SMILES grammar to include this property.

\texttt{ringbond}s are constrained to come in pairs, with the same bond label on both sides.
Whenever a given \texttt{ringbond} is observed, flip a bit in the random access memory corresponding to the ring number (the set of digits after the \texttt{bond}). When the \texttt{ringbond} bit is flipped on, record the associated \texttt{bond} in the random access memory associated with the ring number; when the \texttt{ringbond} bit is flipped off, require that the new \texttt{bond} matches the recorded \texttt{bond}, and clear the random access memory of the \texttt{bond}. The molecule is only allowed to terminate (\texttt{rest\_of\_chain} produces $\epsilon$ rather than \texttt{bond}? \texttt{chain}) when all \texttt{ringbond} bits are off (parity is even). The decoder may receive as input which \texttt{ringbond}s are open, and the associated \texttt{bond} type, so it can preferentially close them.

The set of nested atomic contexts induced by \texttt{chain}, \texttt{branched\_atom}, and \texttt{branch} can be arbitrarily deep, corresponding to the depth of branching in the spanning tree realized by a SMILES string. As a result, the set of SMILES symbols describing bonds to a single atom can be arbitrarily far away from t=he associated \texttt{atom}. However, once a branch is entered, it must be traversed in its entirety before the SMILES string can return to the parent atom. For each atom, it is sufficient to push the valence shell information onto the stack as it is encountered. If the SMILES string enters a branch while processing an atom, simply push on a new context, with a new associated root atom. Once the branch is completed, pop this context off the stack, and return to the original atom.

More specifically, each atom in the molecule is completely described by a single \texttt{branched\_atom} and the \texttt{bond} preceding it (from the \texttt{rest\_of\_chain} that produced the \texttt{branched\_atom}). Within each successive pair of \texttt{bond} and \texttt{branched\_atom}, track the sum of the incoming \texttt{rest\_of\_chain}\texttt{bond}, the internal \texttt{ringbond}  and \texttt{branch} \texttt{bond}s, and outgoing \texttt{rest\_of\_chain} \texttt{bond} (from the succeeding \texttt{rest\_of\_chain}) on the stack. That is, each time a new bond is observed from the atom, pop off the old valence shell count and push on the updated count. Require that the total be less than a bound set by the atom; any remaining bonds are filled by implicity hydrogen atoms. Provide the number of available bonds as input to the decoder RNN, and mask additional \texttt{ringbond}s and \texttt{branch}es once the number of remaining available bonds reaches one (if there are still open \texttt{ringbond}s) or zero (if all \texttt{ringbond}s are closed). Mask the outgoing \texttt{bond}, or require that \texttt{rest\_of\_chain} produce $\epsilon$, based upon the number of remaining available bonds.

\subsection{Redundancy in graph-based and SMILES representations of molecules} \label{sec:graph_redundancy}

To avoid the degeneracy of SMILES strings, for which there are many encodings of each molecule, some authors have advocated the use of graph-based representations~\cite{li2018learning, liu2018constrained, ma2018constrained, simonovsky2018graphvae}. While graph-based processing may produce a unique representation in the encoder, it is not possible to avoid degeneracy in the decoder. Parse trees~\cite{dai2018syntax, kusner2017grammar}, junction trees~\cite{jin2018junction}, lists of nodes and edges~\cite{li2018learning, liu2018constrained, samanta2018nevae}, and vectors/matrices of node/edge labels~\cite{de2018molgan, ma2018constrained, simonovsky2018graphvae} all imply an ordering amongst the nodes and edges, with many orderings describing the same graph. Canonical orderings can be defined, but unless they are obvious to the decoder, they make generative modeling harder rather than easier, since the decoder must learn the canonical ordering rules. 
Graph matching procedures can ensure that probability within a generative model is assigned to the correct molecule, regardless of the order produced by the decoder~\cite{simonovsky2018graphvae}. However, they do not eliminate the degeneracy in the decoder's output, and the generative loss function remains highly multimodal. 

\end{appendix}
\end{document}